\documentclass[10pt,logo,copyright]{nvidiatechreport}
\renewcommand{\today}{November 2026}

\usepackage[numbers,sort]{natbib}
\usepackage{multirow}
\usepackage{multicol}
\usepackage{hhline}
\usepackage{gensymb}
\usepackage{subcaption}
\usepackage{wrapfig}
\usepackage{makecell}
\usepackage{placeins}
\usepackage[nameinlink]{cleveref}

\crefname{figure}{Fig.}{Figs.}
\Crefname{figure}{Fig.}{Figs.}
\crefname{table}{Tab.}{Tabs.}
\Crefname{table}{Tab.}{Tabs.}
\definecolor{ategreen}{RGB}{118,185,0}
\definecolor{ategrey}{RGB}{118,185,0}
\definecolor{movbest}{rgb}{0.96, 0.57, 0.58}
\definecolor{movsecond}{rgb}{0.98, 0.78, 0.57}

\usepackage{amsmath,amsfonts,bm}

\def\eqref#1{equation~\ref{#1}}

\def\1{\bm{1}}

\newcommand{\ourmethod}{\textsc{MetricTrail}}

\DeclareMathAlphabet{\mathsfit}{\encodingdefault}{\sfdefault}{m}{sl}
\SetMathAlphabet{\mathsfit}{bold}{\encodingdefault}{\sfdefault}{bx}{n}

\title{\ourmethod{}: Revisiting the Power of Metric Depth and Local Optimization for Long-Sequence Streaming 3D Reconstruction}
\author{Jingnan Gao\textsuperscript{1,2}, Jiahui Huang\textsuperscript{1}, Sven Elflein\textsuperscript{1,3}, Jianhao Zheng\textsuperscript{1,4}, Bingxin Ke\textsuperscript{1}, Xiaokang Yang\textsuperscript{2},\authorcr
Yichao Yan\textsuperscript{2}, Laura Leal-Taix\'{e}\textsuperscript{1}, Qunjie Zhou\textsuperscript{1}, Jiawei Ren\textsuperscript{1,\ensuremath{\dagger}}, Shengyu Huang\textsuperscript{1,\ensuremath{\dagger}}\\
\textsuperscript{1}NVIDIA \quad \textsuperscript{2}Shanghai Jiao Tong University \quad \textsuperscript{3}University of Toronto, Vector Institute
\textsuperscript{4}Stanford University\\
\textsuperscript{\ensuremath{\dagger}}Equal advising  
\url{https://research.nvidia.com/labs/sil/projects/metrictrail/}
}

\hypersetup{
    pdftitle={MetricTrail: Revisiting the Power of Metric Depth and Local Optimization for Long-Sequence Streaming 3D Reconstruction},
    pdfauthor={Jingnan Gao, Jiahui Huang, Sven Elflein, Jianhao Zheng, Bingxin Ke, Xiaokang Yang, Yichao Yan, Laura Leal-Taix\'{e}, Qunjie Zhou, Jiawei Ren, Shengyu Huang}
}

\begin{document}
\maketitle
\vspace{-\baselineskip}

\begin{abstract}
We present \textbf{\ourmethod{}}, a streaming system for accurate camera
motion estimation and dense reconstruction from long monocular videos.
To keep inference efficient over long sequences, we revisit DPVO's sparse
local estimator. We find that single-image metric depth substantially
improves its trajectory accuracy, without retraining the correspondence
network or enlarging the optimization window. Metric predictions initialize
patch depths and remain as soft constraints during optimization. We further
improve correspondence estimation through learned adapters that augment
matching features and recurrent context with frozen DINOv3 representations.
The correspondence estimator and adapters are trained on diverse short clips
without additional long-sequence training. This modular design also allows
the depth provider to be independently upgraded and supports supplied camera
intrinsics and calibrated stereo observations without retraining. Under the
reported RGB-only protocols, \ourmethod{} achieves the lowest average
trajectory error among evaluated methods on KITTI, Oxford Spires, and VBR.
Pairing the estimated poses with single-image depth predictions also produces
strong dense reconstruction on indoor and outdoor benchmarks and coherent
reconstructions of in-the-wild videos. With supplied intrinsics, the system
processes KITTI-02 at 12.70 FPS with 3.21 GB peak GPU memory.
\end{abstract}

{\hbadness=10000\abscontent\par}

\begin{center}
\captionsetup{type=figure}
\includegraphics[width=\textwidth]{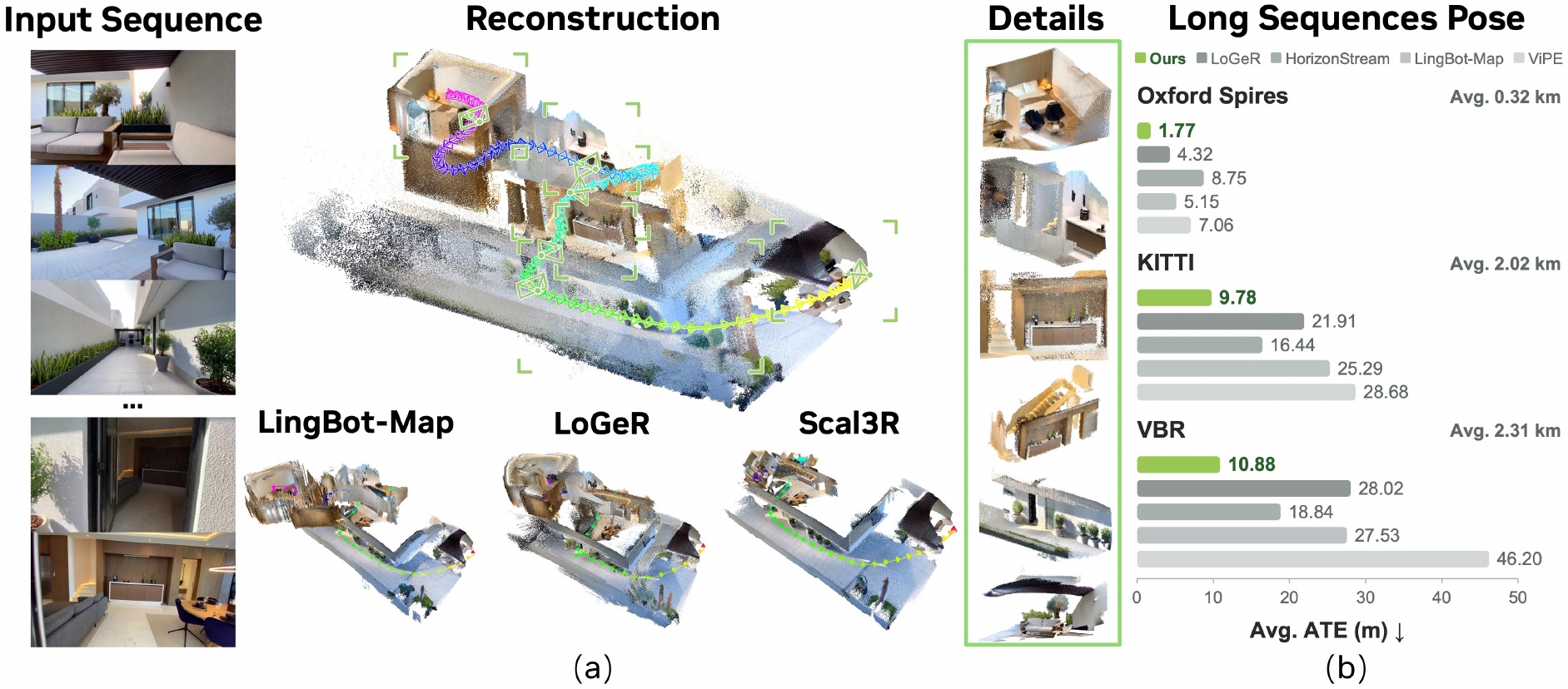}
\captionof{figure}{\textbf{\ourmethod{} reconstructs long videos with coherent scene geometry and accurate camera trajectories.}
(a) Across an indoor--outdoor transition, our reconstruction preserves global
layout and fine local structure, while prior methods show
deformation and misalignment. (b) We achieve lower average trajectory error
than the baselines on three long-sequence benchmarks.}
\label{fig:teaser}
\end{center}

\section{Introduction}
\label{sec:introduction}

Streaming 3D reconstruction turns thousands of monocular observations
into a coherent geometric model of the world. For robotic navigation and
large-scale mapping, this model must remain consistent as the camera moves
through new environments and revisits familiar places. Over long trajectories,
camera-motion and scale errors can accumulate, misaligning observations of the
same scene. Maintaining consistency with manageable computation and memory
is a central challenge for long-video reconstruction.

Recent systems seek long-sequence consistency through learned temporal
context and explicit geometric refinement. CUT3R~\citep{cut3r} maintains a
persistent recurrent state, but accumulated drift remains a challenge on very
long sequences. LingBot-Map~\citep{chen2026lingbotmap} structures attention
memory to retain geometric information across time; for extremely long
inputs, its windowed mode introduces alignment errors that can accumulate
across window boundaries. Optimization-based camera estimation provides a
complementary way to enforce multi-view consistency. ViPE~\citep{huang2025vipe} combines learned
correspondences and metric-depth regularization with local tracking and global
bundle adjustment (BA). However, its original pipeline performs global BA
over all retained frames, including a final refinement after processing the
full video, increasing memory and computation as the sequence grows.

These costs motivate strengthening local estimation with more accurate
depth and metric scale cues, which modern single-image depth models can
now provide.
We therefore revisit DPVO~\citep{teed2023dpvo}, which alternates sparse patch
matching with local BA within a bounded active window. Although efficient,
the original estimator falls substantially behind recent reconstruction
systems on our long-sequence benchmarks. This raises a question:
\emph{How can we improve long-sequence reconstruction without enlarging
the optimization window?}

We introduce \textbf{\ourmethod{}}, a streaming reconstruction system that
strengthens DPVO with single-image metric depth and further improves
correspondence estimation with visual foundation representations.
In the original estimator, new patch depths are initialized heuristically
and subsequently refined through reprojection constraints alone. Building on
metric-depth regularization in learned odometry~\citep{cheng2024romeo,huang2025vipe},
we initialize patch depths with MoGe-2~\citep{wang2025moge2} predictions
and retain them as robust soft constraints during optimization. These
predictions provide both geometry specific to each image location and a
learned reference for physical scale, rather than simply propagating the
estimator's current scale as the camera advances. We find that these
single-image metric priors alone substantially improve the long-sequence
accuracy of the pretrained sparse estimator, without retraining its
correspondence network or enlarging its optimization window.

Metric depth helps predict where a patch should appear, but repeated
structures, weak texture, and appearance changes can still make its
correspondence ambiguous. We therefore augment native matching features and
recurrent context with frozen DINOv3 representations~\citep{simeoni2025dinov3}
through learned adapters. These representations provide complementary
appearance and contextual cues for predicting correspondence targets and
confidence weights. The correspondence estimator and adapters are trained on
diverse public datasets using short clips, preserving the sparse patch graph
and local solver without additional long-sequence training. Metric predictions
and constraints are not used during training and enter only at inference.

\ourmethod{} achieves the lowest average full-sequence absolute trajectory
error (ATE) among the evaluated methods on KITTI, Oxford Spires, and VBR
under the reported RGB-only protocols, while retaining efficient local
inference. Its dense reconstructions also achieve the lowest Chamfer distance
on 7Scenes and Oxford Spires using single-image predictions without dense
depth optimization. On challenging in-the-wild video (\cref{fig:teaser}), it
preserves coherent layout and fine local detail across large indoor--outdoor
depth changes. Beyond monocular RGB, it supports supplied intrinsics
and calibrated stereo without retraining, and allows independent upgrades
to its depth provider.

Our contributions are threefold:
\begin{itemize}[
    leftmargin=1.2em,
    rightmargin=0pt,
    labelsep=0.35em,
    topsep=0pt,
    itemsep=2pt,
    parsep=0pt,
    partopsep=0pt
]
    \item \textbf{Long-sequence streaming reconstruction system.}
    We introduce \ourmethod{}, a streaming system that uses metric depth
    and pretrained visual representations to reconstruct long videos through
    efficient sparse geometric estimation.
    \item \textbf{Sparse local estimator.}
    Metric-depth initialization and persistent constraints improve long-sequence
    accuracy without correspondence-network retraining or a larger optimization window.
    \item \textbf{Long-sequence accuracy and efficiency.}
    We demonstrate leading trajectory accuracy under the evaluated
    protocols, strong dense reconstruction, and efficient local inference
    without additional long-sequence training.
\end{itemize}

\section{Related Work}
\label{sec:related_work}

\noindent\textbf{Optimization-based 3D Reconstruction.}
Optimization-based reconstruction optimizes explicit camera and scene variables
for multi-view consistency. Classical systems
solve offline SfM~\citep{colmap,glomap} or maintain online
estimates~\citep{campos2021orbslam3,dso}. Local windows bound update costs;
global refinement and loop closure address errors accumulated beyond them.
Learning augments these solvers with geometric
priors and feature-metric residuals~\citep{yang2020d3vo,banet}.
DROID-SLAM~\citep{teed2021droid} and DPVO learn recurrent
correspondence updates within differentiable BA; other
approaches~\citep{azhari2025dinovo,flowmap,wang2023coslam,pataki2026vidmap}
incorporate pretrained features, learned matches, or neural scene
representations. DINO-VO~\citep{chen2026dinovo} combines pretrained features
and adaptive patch selection with sparse BA, initializing patch depths with
Depth Anything V2 predictions rescaled to the odometry scale. Such initialization
differs from retaining a metric reference during subsequent optimization.
RoMeO~\citep{cheng2024romeo} and ViPE retain
metric-depth constraints in local BA; DROID-W~\citep{li2026droidw} also
regularizes depth and uses DINOv2~\citep{oquab2024dinov2} features to estimate dynamic uncertainty.

\noindent\textbf{Feed-Forward 3D Reconstruction.}
Feed-forward models~\citep{dust3r,mast3r,vggt,zhang2025monst3r}
learn to predict geometry directly.
Subsequent work advances unified and metric
prediction~\citep{da3,mapanything,dens3r,liu2026foundationgeo,amb3r}
and model architecture~\citep{more,pi3,dejaview,shen2025fastvggt}.
MoGe~\citep{wang2024moge} and MoGe-2 provide single-image priors;
MoGe-2 recovers metric depth while preserving fine relative geometry.
Long videos require consistent geometry without retaining the full observation history.
Persistent or adapted neural states~\citep{spann3r,cut3r,mem3r,loger,scal3r}
and structured temporal
aggregation~\citep{long3r,stream3r,streamvggt,vigeo,wang2025flashvggt,r3}
preserve context across frames or chunks while managing memory.
LingBot-Map combines anchor context for coordinate
grounding, local pose references, and compact trajectory tokens.
HorizonStream~\citep{cheng2026horizonstream} couples local geometric
attention to bounded recurrence and learned pose/scale readouts, with
optional loop-closure optimization.
Explicit geometric refinement complements retained temporal context by correcting
accumulated pose and structure errors.

\noindent\textbf{Hybrid 3D Reconstruction.}
Hybrid systems refine learned predictions through geometric backends.
VGGSfM~\citep{vggsfm} and MASt3R-SfM~\citep{mast3rsfm} optimize learned tracks
or matches globally, while MASt3R-SLAM~\citep{murai2025mast3rslam} builds an
online estimator from point-map priors.
VGGT-SLAM~\citep{vggtslam} and VGGT-Long~\citep{vggtlong} align local
submaps or overlapping reconstructions; VGGT-SLAM
2.0~\citep{maggio2026vggtslam2} additionally corrects keyframes within and
across submaps.
Marginalized Bundle Adjustment~\citep{mba} refines cameras and predicted
depth maps through per-frame scale and shift adjustments, while
BLASt3R~\citep{leroy2026blast3r} combines learned matching and monocular
depth regularization with BA and periodic global refinement.
We focus on long-sequence accuracy within a fixed local window:
persistent metric-depth constraints help stabilize scale across updates,
while richer visual representations
improve correspondences for joint pose and structure refinement.

\section{Method}
\label{sec:method}

\subsection{Problem Setup and System Overview}
\label{sec:method_overview}

Given a long monocular RGB sequence $\{I_i\}_{i=1}^{T}$, our goal is to
estimate world-to-camera transformations $\{\mathbf{T}_i\in\mathrm{SE}(3)\}_{i=1}^{T}$ and a
dense metric reconstruction in a shared world frame. The camera intrinsics
$\mathbf{K}$ may be supplied or estimated from the RGB stream. We also support
synchronized, rectified stereo; in this setting, $\mathbf{T}_i$ denotes the
left-camera pose and the right camera is related to it by a known, fixed rig
transform.

At the center of \ourmethod{} is DPVO's bounded sparse local
estimator~\citep{teed2023dpvo}. At each refinement step, the current geometric
state determines where the matcher looks, while the image representation
determines how that projection is corrected. We strengthen the former with a
metric depth prior and the latter with visual foundation features, while
retaining DPVO's sparse state and alternating matching--optimization loop. The
resulting metric-scale trajectory then places dense frame-level geometry in a
common coordinate frame to form the final reconstruction.

\begin{figure}[t]
    \centering
    \includegraphics[width=\textwidth]{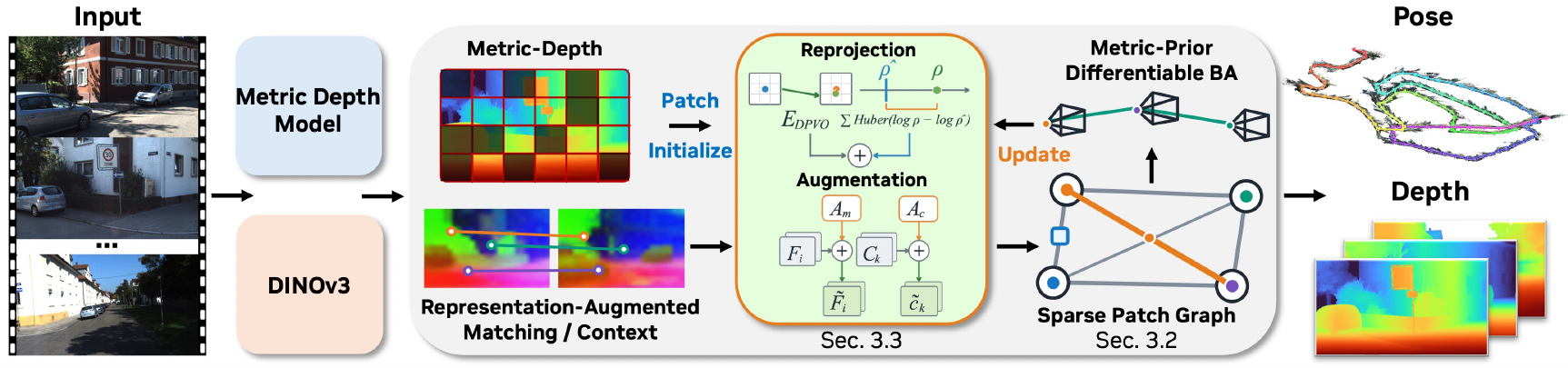}
    \caption{\textbf{System overview.} For each input frame, single-image
    metric depth initializes patch depths and constrains BA, while frozen
    DINOv3 features augment matching and recurrent context through learned
    adapters. Within a bounded active window, patch correspondences drive
    joint pose and depth refinement, and the updated geometry guides the next
    matching step. The resulting poses align dense per-frame depth maps into
    a shared reconstruction.}
    \label{fig:system_overview}
\end{figure}

\subsection{DPVO Recap}
\label{sec:dpvo_recap}

\noindent\textbf{Sparse patch graph.}
DPVO uses a sparse patch graph to organize the visual
correspondences used for joint pose and depth estimation. The graph connects image patches to camera frames: each patch
is anchored in a source frame and tracked across several nearby
target frames. These observations provide geometric constraints
on the camera poses and the patch's depth.

Specifically, each patch $k$ has a fixed center $\mathbf{u}_k$
in its source frame $s(k)$ and an optimizable inverse depth
$\rho_k$. It covers a $3\times3$ region on the downsampled
feature grid, with all nine locations sharing the same depth.
An edge $(k,j)\in\mathcal{E}_{\mathrm{act}}$ connects patch $k$
to target frame $j$. Given the current poses and inverse depth,
the projected patch center is
\begin{equation}
    \mathbf{q}_{kj}(\mathbf{T}_{j}, \mathbf{T}_{s(k)}, \rho_k)
    =
    \pi_{\mathbf{K}}\!\left(
    \mathbf{T}_j\mathbf{T}_{s(k)}^{-1}
    \pi_{\mathbf{K}}^{-1}(\mathbf{u}_k,\rho_k)
    \right),
    \label{eq:dpvo_projection}
\end{equation}
where $\pi_{\mathbf{K}}^{-1}$ back-projects the source pixel
at depth $1/\rho_k$ into 3D, and $\pi_{\mathbf{K}}$ projects
the transformed point into the target frame.

\noindent\textbf{Matching and geometric refinement.}
DPVO alternates between learned patch matching and differentiable
BA. The matching step predicts where each
patch should appear in its target frames, while BA jointly
refines the camera poses and patch depths to agree with these
predictions.

DPVO encodes each image $I_i$ into a matching feature map
$\mathbf{F}_i$ and a context feature map $\mathbf{C}_i$.
At iteration $t$, it correlates the source patch features with
local neighborhoods in $\mathbf{F}_j$ around the projected
patch, measuring visual similarity near the current geometric
estimate. A recurrent update operator combines these correlations
with source context features and information retained from
earlier iterations, while exchanging information across related
graph edges. For each edge $(k,j)$, it predicts a 2D correction
$\boldsymbol{\Delta}_{kj}^{(t)}$ to the projected patch center
and confidence weights $\mathbf{w}_{kj}^{(t)}\in(0,1)^2$.
The correspondence target passed to BA is
$\widehat{\mathbf{q}}_{kj}^{(t)}
=\mathbf{q}_{kj}^{(t)}+\boldsymbol{\Delta}_{kj}^{(t)}$,
where $\mathbf{q}_{kj}^{(t)}$ is the patch center projected
using the current poses and inverse depth.

The BA step holds these correspondence targets and confidence weights fixed
and jointly optimizes the active poses and patch inverse depths by
minimizing the weighted reprojection error:
\begin{equation}
    E_{\mathrm{DPVO}}(\mathbf{T},\boldsymbol{\rho})
    =
    \sum_{(k,j)\in\mathcal{E}_{\mathrm{act}}}
    \left\|
        \mathbf{q}_{kj}(\mathbf{T}_{j}, \mathbf{T}_{s(k)}, \rho_k)
        -\widehat{\mathbf{q}}_{kj}^{(t)}
    \right\|^2_{\operatorname{diag}(\mathbf{w}_{kj}^{(t)})}.
    \label{eq:dpvo_ba}
\end{equation}
The two components of $\mathbf{w}_{kj}^{(t)}$ weight the horizontal and vertical
reprojection residuals, respectively. After BA, the updated poses and inverse depths determine new patch
projections, which are used to compute the correlations
for the next recurrent iteration.
This loop operates within
a bounded active window: new frames enter, redundant frames are removed, and
older poses remain in the output trajectory after leaving the optimizer.

\subsection{Foundation Priors for Geometry and Matching}
\label{sec:prior_augmentation}
\paragraph{Metric Depth for Sparse Geometry.}
\label{sec:metric_prior}

During bootstrap, DPVO seeds patch inverse
depths randomly; once initialized, it assigns new patches the median inverse
depth of patches from the preceding three active frames. The median is a
stable numerical seed, but it contains no geometry specific to the new patch
and carries forward the current monocular scale rather than correcting it.

To provide a scene-specific starting point, we initialize each patch from a
metric image-depth prediction. Given the camera intrinsics, frozen
MoGe-2 independently predicts a metric $z$-depth map
$\widehat{\mathbf{D}}_i$ for each frame using the corresponding field of view.
After aligning it with DPVO's image coordinates, we initialize a valid patch
with $\rho_k^{(0)}\leftarrow\widehat{\rho}_k=
1/\widehat{\mathbf{D}}_{s(k)}(\mathbf{u}_k)$. The first reprojection therefore
starts from a depth tailored to that image location and expressed in metric
units.

Initialization alone cannot prevent later BA updates from drifting away from
the predicted metric scale. We therefore retain the same observation as a
robust unary depth factor. For the set $\mathcal{M}$ of active patches with
valid metric predictions, we optimize
\begin{equation}
    E_{\mathrm{ours}}(\mathbf{T},\boldsymbol{\rho})
    =E_{\mathrm{DPVO}}(\mathbf{T},\boldsymbol{\rho})
    +\lambda_{\mathrm{m}}
    \sum_{k\in\mathcal{M}}
    \operatorname{Huber}_{\delta}
    \!\left(\log\rho_k-\log\widehat{\rho}_k\right).
    \label{eq:metric_ba}
\end{equation}
The optimized patch depth remains free to reconcile the monocular prediction
with multi-view evidence. After an initial stabilization period, we activate
this factor and disable DPVO's scale normalization. The metric observations
then define the reconstruction scale, while the log-domain Huber penalty lets
multi-view evidence override unreliable predictions.

\paragraph{Robust Semantic Priors for Matching.}
\label{sec:visual_prior}

The metric prior improves the geometry from which local search begins, but
choosing among nearby candidates remains a visual matching problem.
Changes in viewpoint and appearance can make these candidates difficult to
distinguish, particularly outside the odometry network's training distribution.
We therefore supplement its task-specific features with representations learned
from large-scale visual pretraining, using a frozen DINOv3-S/16
encoder. Let
$\mathbf{Z}_i=\mathcal{F}_{\mathrm{DINO}}(I_i)$ denote its spatial feature map.
Learned residual adapters align these features with DPVO's spatial grid and
project them into its matching and context dimensions. The matching branch
augments the full feature map, while the context branch conditions each source
patch's recurrent update:
\begin{equation}
    \widetilde{\mathbf{F}}_i
    =\mathbf{F}_i+\mathcal{A}_{\mathrm{m}}(\mathbf{Z}_i),
    \qquad
    \widetilde{\mathbf{c}}_k
    =\mathbf{C}_{s(k)}(\mathbf{u}_k)
    +\mathcal{A}_{\mathrm{c}}(\mathbf{Z}_{s(k)},\mathbf{u}_k).
    \label{eq:dino_fusion}
\end{equation}
We sample the source patch from $\widetilde{\mathbf{F}}_{s(k)}$ and correlate it
against $\widetilde{\mathbf{F}}_j$. The context adapter samples its aligned
features at $\mathbf{u}_k$ before the final projection and addition, producing
the patch context $\widetilde{\mathbf{c}}_k$ used by the same updater. DINOv3
remains frozen, and zero-initializing the adapters' output projections ensures
that the fused model initially reproduces the pretrained DPVO predictions. The adapters
then learn how to use the foundation features without changing the sparse
graph or geometric solver.

\paragraph{Training the Local Estimator.}
\label{sec:method_training}

The two priors play different roles during training and inference.
MoGe-2 enters only through inference-time initialization and BA factors, so
the correspondence network never sees metric predictions during training.
DINOv3, by contrast, changes the matcher inputs during adaptation. Its encoder
remains frozen; starting from the official DPVO checkpoint, we optimize the
DPVO modules and residual adapters with the original pose and induced
patch-motion supervision.

The network learns local correspondence corrections, while repeated geometric
optimization composes these measurements over time. This division allows us
to train on diverse short clips without requiring long-trajectory supervision. Each
sample contains 15 frames drawn from TartanAir~\citep{wang2020tartanair},
TartanGround~\citep{tartanairv2}, CarlaOcc~\citep{feng2026carlaocc},
LightwheelOcc~\citep{lightwheel2024}, or
OmniWorld-Game~\citep{zhou2026omniworld}. No evaluation sequence or dedicated
long-trajectory pose corpus is used.

\subsection{Complete Long-Video System}
\label{sec:system_integration}

Compared with learned streaming reconstructors such as
HorizonStream and
LingBot-Map, our optimization-based estimator can
directly incorporate additional observations and geometric constraints at
inference time, without retraining.

\noindent\textbf{Camera calibration.} Camera intrinsics $\mathbf{K}$ are supplied or estimated once from an RGB
prefix. Following ViPE~\citep{huang2025vipe}, MoGe-2 initializes a shared
camera model, and a DROID-style backend~\citep{teed2021droid} jointly refines
its focal correction, auxiliary poses, and depths under metric-depth
constraints. We then fix $\mathbf{K}$, discard the auxiliary state, and
replay the prefix through the local estimator before continuing online.
Revisit-based focal refinement remains an optional retrospective step.

\noindent\textbf{Stereo input.} For synchronized, rectified stereo, each left-image patch is additionally
matched to the right image~\citep{teed2021droid}. The known rig transform
provides a same-time reprojection constraint on patch geometry, without
introducing right-camera pose variables or requiring stereo-specific training.

\noindent\textbf{Loop closure.} Following VGGT-Long~\citep{vggtlong}, Scal3R~\citep{scal3r}, and
HorizonStream~\citep{cheng2026horizonstream}, we use similar loop closure
based on retrieval and pose-graph optimization. The graph connects
camera-pose nodes with sequential odometry edges.
SALAD~\citep{izquierdo2024salad} retrieves distant near-duplicate frames;
each pair adds a weak identity relative-pose edge, softly encouraging the
poses to coincide. Optimizing poses against both edge types corrects drift
while retaining metric scale.

\noindent\textbf{Dense reconstruction.} We back-project per-frame metric depth maps into the world frame using
$\mathbf{T}_i^{-1}$. BA refines camera poses and sparse patch depths,
while the dense predictions remain fixed. We can therefore reassemble the
dense reconstruction using updated poses, including those from optional
pose-graph correction, without rerunning local estimation.

\section{Experiments}
\label{sec:experiments}


\definecolor{movbest}{RGB}{118,185,0}
\definecolor{movsecond}{RGB}{220,241,182}
\newcommand{\movbestcell}[1]{\cellcolor{movbest}{#1}}
\newcommand{\movsecondcell}[1]{\cellcolor{movsecond}{#1}}

\subsection{Experimental Setup}
\label{sec:experimental_setup}

\noindent\textbf{Evaluation Datasets and Metrics.}
\label{sec:evaluation_datasets}
We evaluate long-sequence \textit{camera estimation} on KITTI Odometry
sequences 00--10~\citep{geiger2012kitti}, ten Oxford Spires
sequences~\citep{tao2025spires}, and all seven VBR
sequences~\citep{brizi2024vbr}. Following LingBot-MAP~\citep{chen2026lingbotmap}, we process all three datasets without temporal
subsampling. The Oxford split contains 38,380 frames. We report full-sequence
absolute trajectory error (ATE), measured as RMSE in meters after a single
$\mathrm{Sim}(3)$ alignment per sequence.
For \textit{dense reconstruction}, we evaluate 7Scenes~\citep{shotton2013scene},
TUM-Dynamic~\citep{sturm2012benchmark}, and Oxford Spires. We
report symmetric Chamfer distance (CD) and $\mathrm{F1}_{\tau}$ after
sequence-level $\mathrm{Sim}(3)$ alignment followed by point-to-point ICP,
using $\tau=0.25$ m for indoor scenes and $\tau=4$ m on Oxford Spires.
For \textit{video depth estimation}, we follow $\pi^3$~\citep{pi3} on 13 KITTI
videos containing 1,269 frames, reporting Abs Rel and $\delta_1$ accuracy
($\delta<1.25$) after fitting one scale per video without a shift.

\noindent\textbf{Settings and Baselines.}
\emph{Calibrated} uses the provided camera intrinsics, whereas
\emph{Uncalibrated} estimates a shared camera model for each sequence
from RGB alone. Our dense reconstruction results use the uncalibrated setting.
Loop closure is disabled by default; results with loop closure are marked
\emph{w/ LC}. On KITTI, \emph{Stereo} incorporates
synchronized, rectified left--right cameras and the known stereo
baseline into local BA. We compare with optimization-based and hybrid systems,
offline or chunk-based long-sequence methods, and streaming models including
TTT3R~\citep{chen2025ttt3r}, InfiniteVGGT~\citep{yuan2026infinitevggt},
and LongStream~\citep{cheng2026longstream}.
Tables mark the \colorbox{movbest}{best} and
\colorbox{movsecond}{second-best} values among results; where used, underlining identifies the strongest
average within each baseline family.

\subsection{Long Sequence Camera Trajectory Estimation}
\label{sec:long_sequence_benchmarks}

\begin{figure}
    \centering
    \includegraphics[width=\linewidth]{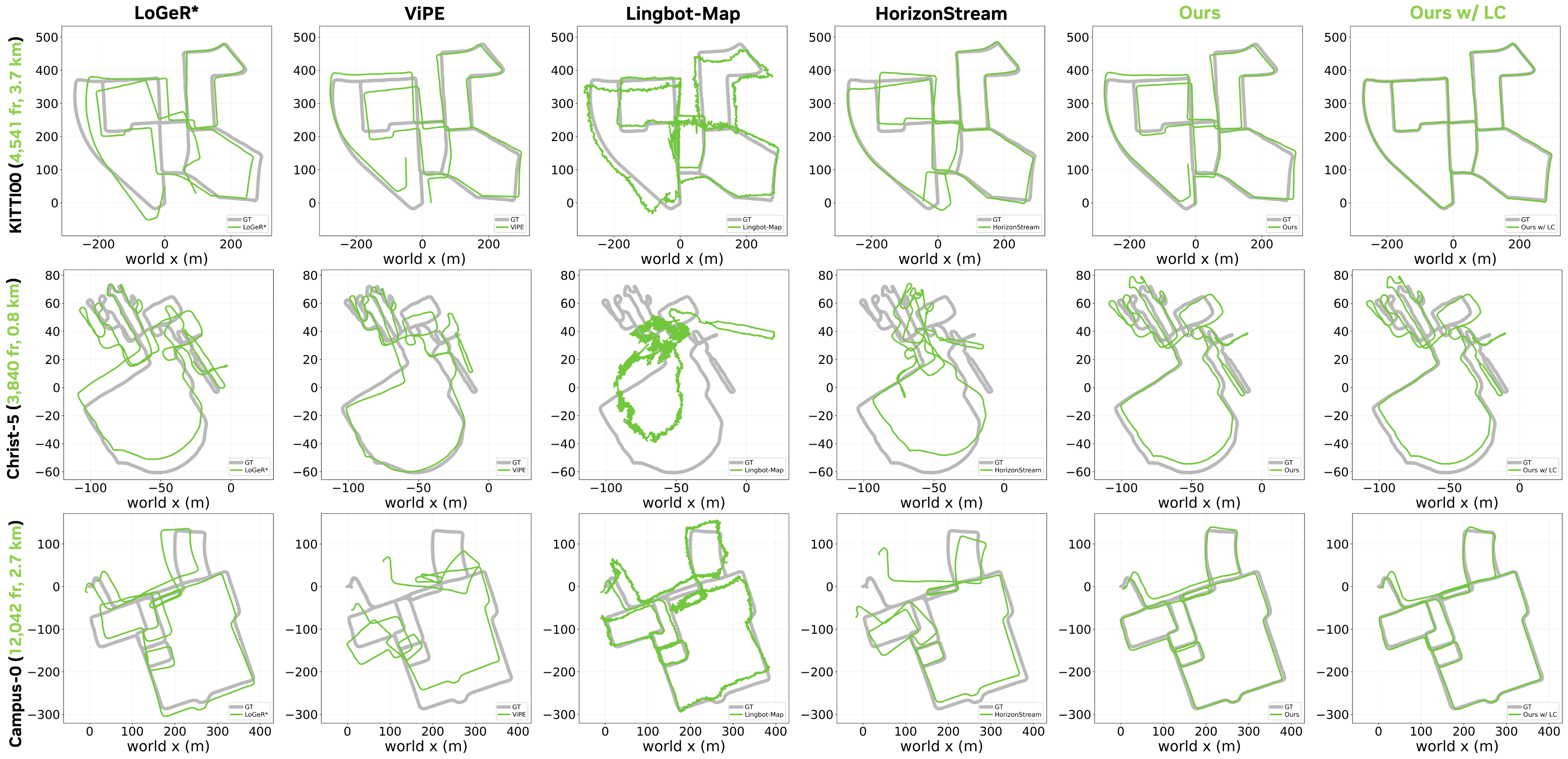}
    \caption{\textbf{Long-sequence camera trajectory estimation.} Ground truth is \textcolor{gray}{gray} and estimates are
    \textcolor{ategreen}{green}.
    Our local estimator reduces deviations; loop closure further improves revisit alignment.}
    \label{fig:trajectory_comparison}
\end{figure}

\begin{table*}[t]
\centering
\caption{\textbf{Full-sequence ATE (m) on KITTI Odometry.}
LC denotes loop closure; LoGeR$^*$ uses alignment optimization.
Underlining marks the best average in each baseline group and our default
RGB-only result, which outperforms all baseline averages.}
\label{tab:kitti_odometry_par}
\scriptsize
\setlength{\tabcolsep}{2.1pt}
\resizebox{\textwidth}{!}{%
\begin{tabular}{cl*{11}{c}c}
\toprule
& \textbf{Method} &
\makecell{\textbf{00}\\4,541 fr.\\3.7 km} &
\makecell{\textbf{01}\\1,101 fr.\\2.5 km} &
\makecell{\textbf{02}\\4,661 fr.\\5.1 km} &
\makecell{\textbf{03}\\801 fr.\\0.6 km} &
\makecell{\textbf{04}\\271 fr.\\0.4 km} &
\makecell{\textbf{05}\\2,761 fr.\\2.2 km} &
\makecell{\textbf{06}\\1,101 fr.\\1.2 km} &
\makecell{\textbf{07}\\1,101 fr.\\0.7 km} &
\makecell{\textbf{08}\\4,071 fr.\\3.2 km} &
\makecell{\textbf{09}\\1,591 fr.\\1.7 km} &
\makecell{\textbf{10}\\1,201 fr.\\0.9 km} & \textbf{Avg.} $\downarrow$ \\
\midrule
\multirow{5}{*}{\rotatebox{90}{\makecell{Optimization\\/ Offline}}}
& DPVO (Calibrated) & 113.11 & \movsecondcell{16.60} & 113.01 & 2.46 & 0.98 & 59.34 & 55.91 & 19.30 & 110.63 & 74.55 & 13.71 & 52.69 \\
& ViPE & 25.87 & 148.34 & 53.35 & 5.29 & 1.95 & 24.22 & 11.72 & \movsecondcell{1.03} & 18.51 & 20.55 & 4.71 & 28.68 \\
& LoGeR & 54.98 & 36.57 & 36.20 & 4.27 & 1.62 & 33.41 & 11.78 & 13.33 & 22.92 & 17.89 & 8.06 & 21.91 \\
& LoGeR$^*$ & 26.19 & 41.26 & 32.21 & 5.02 & 1.62 & 22.65 & 5.49 & 5.04 & 21.96 & 9.03 & 9.44 & 16.35 \\
& Scal3R w/ LC & 4.30 & 45.29 & 42.06 & 3.36 & 1.74 & \movsecondcell{3.30} & \movsecondcell{2.49} & 2.03 & 36.69 & 12.32 & 6.46 & \underline{14.55} \\
\midrule
\multirow{4}{*}{\rotatebox{90}{Online}}
& LongStream & 92.55 & 46.01 & 134.70 & 3.81 & 1.95 & 84.69 & 23.12 & 14.93 & 62.07 & 85.61 & 21.48 & 51.90 \\
& LingBot-MAP & 30.80 & 64.74 & 82.29 & 2.49 & 0.85 & 16.55 & 6.27 & 8.92 & 39.32 & 17.99 & 7.96 & 25.29 \\
& HorizonStream & 26.40 & 20.62 & 84.62 & 5.15 & 0.62 & 12.82 & 4.59 & 5.49 & 19.49 & 25.73 & 11.71 & 19.75 \\
& HorizonStream w/ LC & 13.91 & 20.62 & 69.43 & 5.15 & 0.62 & 6.86 & 6.50 & 2.67 & 19.49 & 23.86 & 11.71 & \underline{16.44} \\
\midrule\midrule
\multirow{4}{*}{\rotatebox{90}{Ours}}
& Ours (Uncalibrated) & 15.29 & 35.94 & 24.69 & \movsecondcell{1.48} & 0.69 & 18.43 & 9.14 & 3.81 & 22.32 & 14.50 & 8.21 & \underline{\textbf{14.05}} \\
& Ours (Uncalibrated) w/ LC & \movsecondcell{2.75} & 35.94 & 18.36 & \movsecondcell{1.48} & 0.69 & 4.44 & 5.61 & 2.10 & 22.32 & 5.64 & 8.21 & 9.78 \\
& Ours (Calibrated) w/ LC & 2.92 & 33.21 & \movbestcell{6.02} & 2.71 & \movsecondcell{0.59} & 3.32 & 9.49 & 2.48 & \movsecondcell{11.94} & \movsecondcell{4.69} & \movsecondcell{3.95} & \movsecondcell{7.39} \\
& Ours (Stereo) w/ LC & \movbestcell{1.43} & \movbestcell{6.33} & \movsecondcell{12.62} & \movbestcell{0.34} & \movbestcell{0.23} & \movbestcell{0.55} & \movbestcell{0.20} & \movbestcell{0.97} & \movbestcell{5.27} & \movbestcell{4.14} & \movbestcell{0.93} & \movbestcell{3.00} \\
\bottomrule
\end{tabular}}
\end{table*}

\noindent\textbf{KITTI Odometry.}
KITTI combines forward vehicle motion with moving traffic and urban
revisits. We also include offline and chunk-based methods to place our
streaming estimator's accuracy in a broader context.
In \cref{tab:kitti_odometry_par}, Scal3R and
HorizonStream, both with loop closure, achieve
the lowest average ATEs in their respective baseline groups: 14.55 m and
16.44 m. Our uncalibrated monocular estimator achieves 14.05 m without loop
closure, outperforming all listed baselines. Providing camera intrinsics
reduces its ATE to 9.10 m. Loop closure improves these
two settings to 9.78 m and 7.39 m, respectively. With calibrated stereo,
ATE falls to 3.66 m without loop closure and 3.00 m with it, confirming
complementary gains from stereo and revisit correction.

\noindent\textbf{Oxford Spires.}
Oxford Spires includes handheld motion with repeated architecture and
viewpoint changes that challenge visual association. Our uncalibrated local
estimator achieves 1.77 m average ATE (\cref{tab:oxford_vbr_avg}), reducing
error by 65.6\% over the strongest online baseline,
LingBot-MAP (5.15 m), and by 58.6\% over the
strongest offline baseline, LoGeR$^*$ (4.28 m).

\begin{table}[t]
\centering
\caption{\textbf{Average full-sequence ATE (m $\downarrow$) on
\textbf{Oxford Spires} and \textbf{VBR}.}}
\label{tab:oxford_vbr_avg}
\footnotesize
\setlength{\tabcolsep}{3pt}
\renewcommand{\arraystretch}{1.15}
\resizebox{\textwidth}{!}{%
\begin{tabular}{@{}c*{14}{c}@{}}
\toprule
& \multicolumn{6}{c}{\textbf{Offline/Chunk}}
& \multicolumn{6}{c}{\textbf{Online}}
& \multicolumn{2}{c}{\textbf{Ours}} \\
\cmidrule(lr){2-7}
\cmidrule(lr){8-13}
\cmidrule(l){14-15}
\textbf{Dataset}
& \makecell{VGGT-\\SLAM}
& \makecell{MASt3R-\\SLAM}
& \makecell{VGGT-\\Long}
& ViPE
& LoGeR
& LoGeR$^*$
& \makecell{Infinite\\VGGT}
& \makecell{Long\\Stream}
& CUT3R
& TTT3R
& \makecell{Horizon\\Stream}
& \makecell{LingBot-\\Map}
& w/o LC
& w/ LC \\
\midrule
Oxford Spires
& 23.42
& 12.12
& 7.41
& 7.06
& 4.32
& \movsecondcell{4.28}
& 31.76
& 15.92
& 14.74
& 9.40
& 8.75
& 5.15
& \movbestcell{1.77}
& \movbestcell{1.77} \\
VBR
& 78.20
& 92.64
& 91.23
& 46.20
& 28.02
& 25.24
& 102.90
& 77.93
& 66.25
& 64.99
& 25.30
& 27.53
& \movsecondcell{16.36}
& \movbestcell{10.88} \\
\bottomrule
\end{tabular}%
}
\end{table}

\noindent\textbf{VBR.}
VBR contains urban routes of 8,815--18,846 frames and up to 5.20 km,
with varying revisit opportunities. \cref{tab:oxford_vbr_avg} summarizes
the results. LoGeR$^*$ and
HorizonStream achieve similar average ATEs of
25.24 m and 25.30 m, respectively. Our uncalibrated local estimator achieves
16.36 m with focal refinement, or 10.88 m with additional loop closure.
Representative trajectories are shown in \cref{fig:trajectory_comparison}.

\subsection{Dense Reconstruction and Depth}
\label{sec:dense_reconstruction}
\begin{figure}[!htbp]
    \centering
    \includegraphics[width=\linewidth]{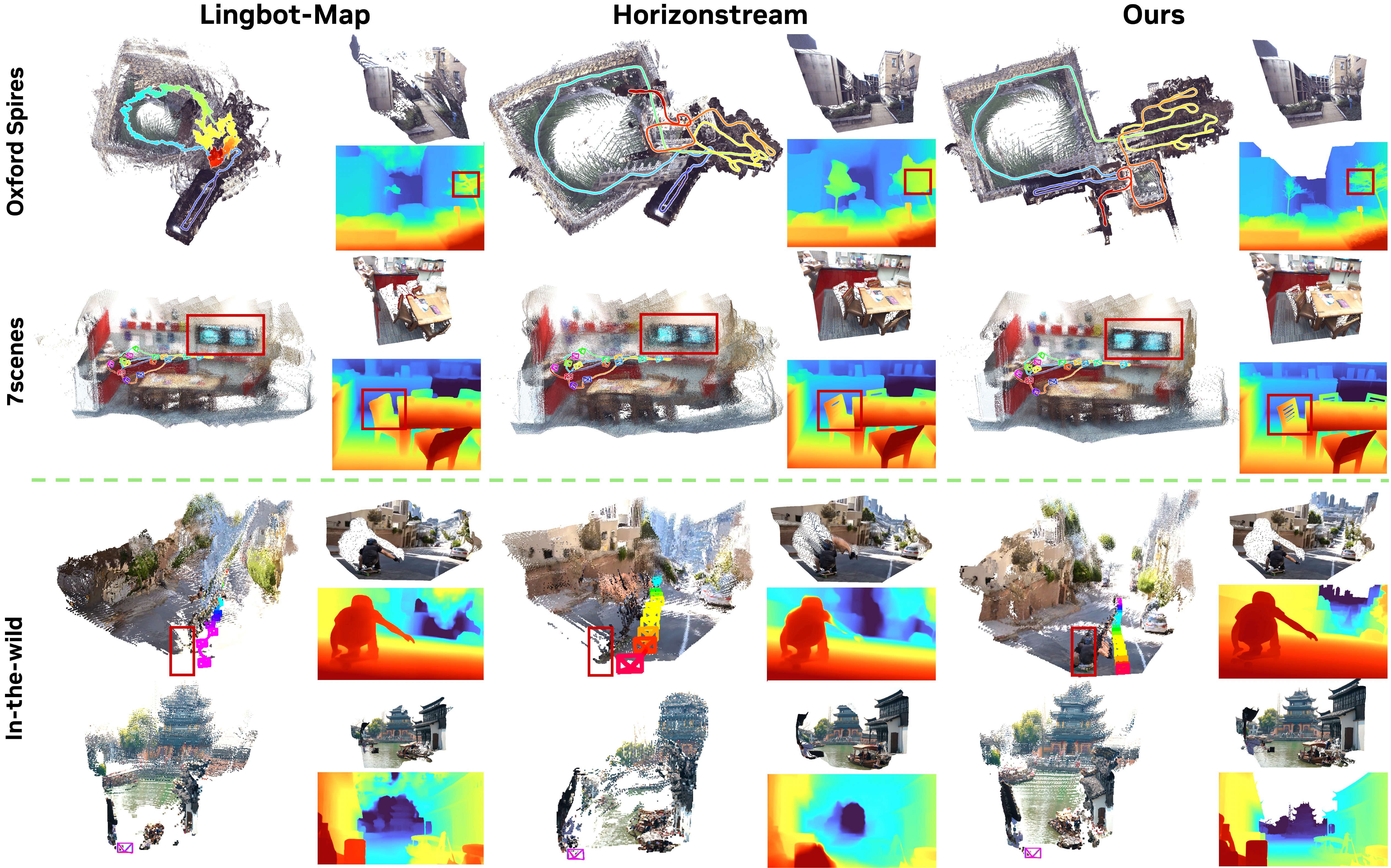}
    \caption{\textbf{Qualitative dense reconstruction comparison.}
    Our method produces more accurate reconstructions with finer point-cloud
    detail. The highlighted regions show clearer structural boundaries and
    more faithful local geometry, alongside detailed depth predictions.}
    \label{fig:dense_qualitative}
\end{figure}

\begin{table}[!t]
\centering
\begin{minipage}[t]{0.6\textwidth}
\centering
\captionsetup{font=small,justification=raggedright,
  singlelinecheck=false}
\captionof{table}{\textbf{Quantitative comparison on dense reconstruction.}
We report Chamfer distance (CD, m; $\downarrow$) and F1 (\%; $\uparrow$).
Our results use RGB-estimated intrinsics without loop closure.}
\label{tab:dense_reconstruction}
\scriptsize
\setlength{\tabcolsep}{2.6pt}
\resizebox{\linewidth}{!}{%
\begin{tabular}{l*{6}{c}}
\toprule
& \multicolumn{2}{c}{\makecell{\textbf{7Scenes}\\($\tau=0.25$ m)}}
& \multicolumn{2}{c}{\makecell{\textbf{TUM-Dynamic}\\($\tau=0.25$ m)}}
& \multicolumn{2}{c}{\makecell{\textbf{Oxford Spires}\\($\tau=4$ m)}} \\
\cmidrule(lr){2-3}\cmidrule(lr){4-5}\cmidrule(lr){6-7}
\textbf{Method}
& \textbf{CD} $\downarrow$ & \textbf{F1} $\uparrow$
& \textbf{CD} $\downarrow$ & \textbf{F1} $\uparrow$
& \textbf{CD} $\downarrow$ & \textbf{F1} $\uparrow$ \\
\midrule
CUT3R         & 0.18 & 75.18                    & 0.11                    & 90.68                    & 7.26                    & 41.22 \\
TTT3R         & 0.11 & 86.98                    & \movbestcell{0.06}      & \movsecondcell{94.45}   & 7.77                    & 40.42 \\
Stream3R-w    & 0.12 & 87.57                    & 0.20                    & 87.40                    & 6.44                    & 47.75 \\
InfiniteVGGT  & 0.08 & 91.96                    & 0.10                    & 92.94                    & 7.38                    & 47.83 \\
LongStream    & 0.06 & 94.60                    & 0.14                    & 88.29                    & 5.69                    & 53.84 \\
LingBot-MAP   & \movsecondcell{0.05} & 96.01 & \movsecondcell{0.08} & \movbestcell{95.97} & \movsecondcell{1.68} & \movsecondcell{90.58} \\
HorizonStream & 0.23 & \movbestcell{98.22}      & 0.12                    & 90.97                    & 2.00                    & 84.69 \\
\midrule\midrule
Ours & \movbestcell{0.04} & \movsecondcell{97.60} & 0.13 & 89.58 & \movbestcell{0.98} & \movbestcell{95.12} \\
\bottomrule
\end{tabular}}
\end{minipage}\hfill
\begin{minipage}[t]{0.36\textwidth}
\centering
\captionsetup{font=small,justification=raggedright,
  singlelinecheck=false}
\captionof{table}{\textbf{Video depth estimation on KITTI.}
We report Abs Rel and $\delta_1$ (\%) with per-video scale alignment.}
\label{tab:kitti_depth}
\scriptsize
\setlength{\tabcolsep}{2.6pt}
\resizebox{\linewidth}{!}{%
\begin{tabular}{@{}lcc@{}}
\toprule
\textbf{Method} & \textbf{Abs Rel} $\downarrow$ &
$\boldsymbol{\delta<1.25}$ $\uparrow$ \\
\midrule
InfiniteVGGT & 0.170 & 78.6 \\
LoGeR & 0.090 & 93.0 \\
LongStream & 0.120 & 87.0 \\
LingBot-MAP & 0.098 & 90.7 \\
HorizonStream & \movsecondcell{0.058} & \movsecondcell{94.8} \\
\midrule\midrule
Ours & \movbestcell{0.057} & \movbestcell{97.8} \\
\bottomrule
\end{tabular}}
\end{minipage}
\end{table}

\cref{tab:dense_reconstruction} summarizes reconstruction accuracy across
indoor, dynamic, and large-scale outdoor environments. On
7Scenes, \ourmethod{} achieves the lowest CD of
0.04 m and the second-highest F1 score of 97.60\%. On Oxford
Spires, it improves over the strongest baseline,
LingBot-MAP, reducing CD from 1.68 to 0.98 m and
increasing F1 from 90.58\% to 95.12\%. Performance is weaker on
TUM-Dynamic, where its 0.13 m CD and 89.58\% F1
trail the leading methods. Thus, the gains on static indoor and large outdoor
scenes do not extend uniformly to scenes with dynamic content.

In \cref{fig:dense_qualitative}, \ourmethod{} preserves the courtyard and
connecting routes on Christ-5, where baseline reconstructions show layout
deformation. The highlighted indoor and in-the-wild regions also show finer
point-cloud detail and more accurate local shape in the depth maps. These
results illustrate the complementary roles of depth and camera motion:
depth captures local geometry, while the optimized trajectory aligns
observations into a coherent scene.
Several long-sequence reconstruction methods exhibit larger depth errors
in \cref{tab:kitti_depth}. We achieve the lowest Abs Rel of 0.057 and
highest $\delta_1$ of 97.8\%. These results use single-image depth predictions
conditioned on calibrated intrinsics shared across each video.

\Needspace{16\baselineskip}
\subsection{Ablation Study}
\label{sec:ablations}

\cref{tab:ablation_summary} shows that metric depth yields the largest
improvement across all three datasets. Multi-dataset adaptation further
improves KITTI and Oxford Spires but increases VBR error; adding DINOv3
improves all three and more than offsets this regression. Broader training
data and pretrained representations thus make complementary contributions.

\begin{figure}[!htbp]
    \centering
    \includegraphics[width=\linewidth]{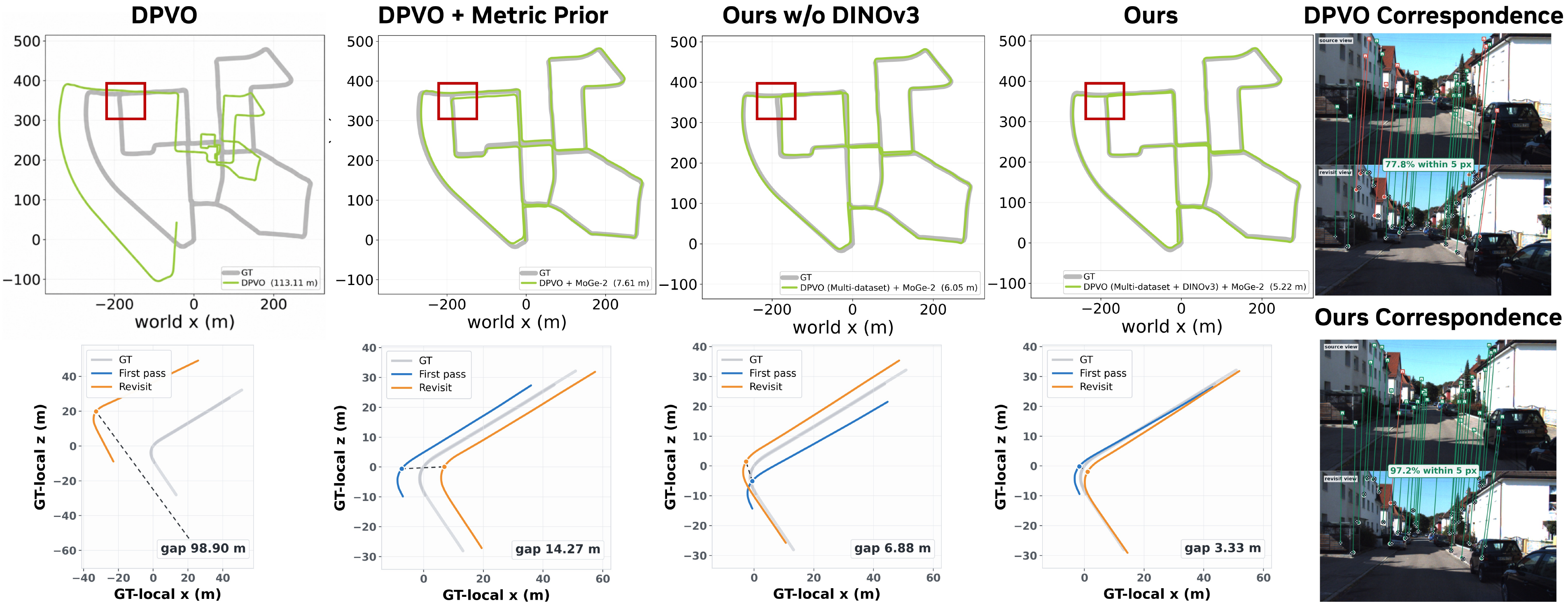}
    \caption{\textbf{Qualitative ablation on KITTI 00.}
    Metric-depth priors substantially reduce accumulated trajectory drift,
    while multi-dataset adaptation and DINOv3 further improve agreement
    between first-pass and revisit trajectories. On the same set of reference
    patches, the full estimator increases the fraction of correspondence
    endpoints within 5 px from $77.8\%$ to $97.2\%$, providing a local diagnostic
    of the stronger visual evidence used by BA.}
    \label{fig:qualitative_ablation}
\end{figure}

\begingroup
\setlength{\intextsep}{0pt}
\begin{wraptable}{r}{0.42\textwidth}
\centering
\captionsetup{font=small,justification=raggedright,
  singlelinecheck=false}
\caption{\textbf{Component ablation.} We progressively add metric depth,
multi-dataset training, and DINOv3 to DPVO and report mean calibrated
ATE (m; $\downarrow$).}
\label{tab:ablation_summary}
\footnotesize
\setlength{\tabcolsep}{1.25pt}
\renewcommand{\arraystretch}{1.02}
\resizebox{\linewidth}{!}{%
\begin{tabular}{@{}
>{\centering\arraybackslash}m{20.5pt}
>{\centering\arraybackslash}m{23.5pt}
>{\centering\arraybackslash}m{25.5pt}@{\hspace{4pt}}
*{3}{>{\centering\arraybackslash}m{27.6pt}}@{}}
\toprule
\multicolumn{3}{c}{\textbf{Components}} &
\multicolumn{3}{c}{\textbf{ATE (m)} $\downarrow$} \\
\cmidrule(lr){1-3}\cmidrule(lr){4-6}
\scriptsize\makecell{\textbf{Metric}\\\textbf{Depth}} & \scriptsize\makecell{\textbf{Multi-}\\\textbf{dataset}} & \scriptsize\textbf{DINOv3} &
\textbf{KITTI} & \textbf{Oxford} & \textbf{VBR} \\
\midrule
-- & -- & -- & 52.69 & 2.59 & 72.25 \\
\textbullet & -- & -- & 12.77 & 1.85 & \movsecondcell{19.57} \\
\textbullet & \textbullet & -- & \movsecondcell{10.32} & \movsecondcell{1.76} & 22.51 \\
\textbullet & \textbullet & \textbullet & \movbestcell{9.10} & \movbestcell{1.71} & \movbestcell{18.90} \\
\bottomrule
\end{tabular}}
\end{wraptable}

\cref{fig:qualitative_ablation} shows that metric-depth initialization and
constraints reduce KITTI 00 ATE from 113.11 to 7.61 m and improve revisit
alignment without frontend retraining. With the same prior, multi-dataset
adaptation and DINOv3 further reduce ATE to 6.05 and 5.22 m.
On the same reference patches, matches within 5 px rise from 77.8\% to
97.2\%, indicating fewer large correspondence errors and more reliable
constraints for local optimization.

\noindent\textbf{Choice of depth provider.}
We evaluate five alternative depth providers in addition to MoGe-2~\cite{wang2025moge2}:
Pi3X~\citep{pi3}, Video Depth Anything (VDA)~\citep{chen2025videodepthanything},
DA3Metric~\citep{da3}, Metric3Dv2~\citep{hu2024metric3dv2}, and
UniDepthV2~\citep{piccinelli2026unidepthv2}. Each provider supplies both the
patch-depth initialization and the persistent metric-depth factors; VDA uses
its metric checkpoint. On KITTI Odometry, Oxford Spires, and VBR, we compare
DPVO, the multi-dataset variant, and the full model with DINO augmentation,
holding the provider fixed across the three variants. As shown in \cref{tab:metric_prior_macro}, the estimator enhancements
\textbf{improve trajectory accuracy across different depth providers},
supporting the \textbf{generality and robustness} of our approach.
\cref{tab:metric_prior_macro} reports their average ATEs alongside the
MoGe-2 results. For the five alternative providers, the full model improves
over DPVO with the same prior in 14 of the 15 prior--dataset combinations,
including all five providers on both KITTI and Oxford Spires. These results
show that the estimator enhancements remain effective across different depth
providers, although the gains depend on the provider and dataset. 
\par
\endgroup

We choose MoGe-2 as the default for two reasons. First, it predicts \textbf{metric
depth and intrinsics from RGB inputs alone}~\citep{wang2025moge2}, allowing one
provider to initialize camera calibration and supply the depth prior.
Although DA3Metric and Metric3Dv2 achieve lower ATE in our calibrated
ablation, their standalone metric-depth inference requires focal-length
information~\citep{da3,hu2024metric3dv2}; using them with RGB-only input
therefore requires a separate source of calibration. Second, depth providers
are shaped by their training distributions, and their relative performance
varies across domains. UniDepthV2, for example, achieves lower ATE than
MoGe-2 on KITTI but higher errors on Oxford Spires and VBR.
MoGe-2 combines broad indoor--outdoor training with \textbf{competitive
results across all three benchmarks}, making it a practical default for a
unified RGB-only system. Our estimator also supports alternative
providers, allowing further gains with suitable calibration.

\begin{table*}[t]
\centering
\captionsetup{font=small,justification=raggedright,singlelinecheck=false}
\caption{\textbf{Metric-prior ablation.} We report average ATE RMSE (m; lower is better).
Each column uses the named metric prior for all three variants; multi-dataset
training and DINO augmentation are added cumulatively. Best and second-best
variants are highlighted within each metric-prior column and dataset.}
\label{tab:metric_prior_macro}
\footnotesize
\setlength{\tabcolsep}{4.0pt}
\renewcommand{\arraystretch}{1.02}
\resizebox{\textwidth}{!}{%
\begin{tabular}{@{}llcccccc@{}}
\toprule
\multirow{2}{*}{\textbf{Dataset}} & \multirow{2}{*}{\textbf{Variant}} & \multicolumn{6}{c}{\textbf{Metric Prior}} \\
\cmidrule(lr){3-8}
 & & \textbf{Pi3X} & \textbf{VDA} & \textbf{DA3Metric} & \textbf{Metric3Dv2} & \textbf{UniDepthV2} &\textbf{MoGe-2}\\
\midrule
\multirow{3}{*}{\textbf{KITTI}} & DPVO & 27.92 & 15.16 & 14.05 & 11.62 & 11.18 & 12.77\\
 & + Multi-dataset & \movsecondcell{24.90} & \movsecondcell{8.97} & \movsecondcell{10.03} & \movsecondcell{10.57} & \movsecondcell{10.80} & \movsecondcell{10.32} \\
 & + DINO (Ours) & \movbestcell{22.35} & \movbestcell{8.88} & \movbestcell{7.97} & \movbestcell{8.21} & \movbestcell{7.69} & \movbestcell{9.10}\\
\midrule
\multirow{3}{*}{\textbf{Oxford Spires}} & DPVO & 5.57 & 2.25 & \movsecondcell{1.28} & 1.42 & \movsecondcell{3.89} & 1.85 \\
 & + Multi-dataset & \movbestcell{4.53} & \movbestcell{1.87} & 1.32 & \movsecondcell{1.33} & 6.04 & \movsecondcell{1.76}\\
 & + DINO (Ours) & \movsecondcell{4.78} & \movsecondcell{1.92} & \movbestcell{1.14} & \movbestcell{1.15} & \movbestcell{3.81} & \movbestcell{1.71} \\
\midrule
\multirow{3}{*}{\textbf{VBR}} & DPVO & \movsecondcell{30.29} & \movsecondcell{24.70} & \movsecondcell{19.17} & \movsecondcell{20.14} & \movbestcell{19.11} & \movsecondcell{19.57}\\
 & + Multi-dataset & 36.13 & 26.61 & 25.85 & 23.58 & 24.61 & 22.51\\
 & + DINO (Ours) & \movbestcell{30.04} & \movbestcell{23.07} & \movbestcell{18.39} & \movbestcell{18.79} & \movsecondcell{19.56} &\movbestcell{18.90} \\
\bottomrule
\end{tabular}}
\end{table*}

\noindent\textbf{Runtime analysis.}
On KITTI-02 with an NVIDIA A100 GPU, \ourmethod{} achieves a steady-state throughput
of 12.70 FPS, compared with 11.37 FPS for LingBot-Map and 9.92 FPS for
HorizonStream. With supplied
intrinsics, time to first output is approximately 0.7 s and peak GPU memory
is 3.21 GB. RGB-only calibration increases start-up latency to approximately
175 s and peak memory to 6.81 GB, while subsequent throughput remains
12.70 FPS.
\section{Conclusion}
\label{sec:conclusion}

We presented \textbf{\ourmethod{}}, a streaming reconstruction system centered
on the finding that single-image metric depth substantially improves sparse
local odometry over long sequences. These gains do not require retraining
the correspondence network or enlarging the optimization window, while
pretrained visual representations further improve correspondence estimation.
The resulting system supports accurate trajectories and strong dense
reconstruction across benchmark and in-the-wild videos without additional
long-sequence training. Its modular design accommodates improved depth
models, supplied camera intrinsics, and calibrated stereo observations.
Together, these results demonstrate that strong image-level priors, coupled
with sparse local optimization, offer an accurate, efficient, and flexible
approach to long-video reconstruction.

\FloatBarrier
{\small
\bibliographystyle{unsrtnat}
\bibliography{reference}
}

\clearpage
\appendix
\section{Appendix}
\label{sec:appendix}
\suppressfloats[t]
\noindent\textbf{Appendix overview.}
\begin{itemize}[leftmargin=0pt,label={},itemsep=2pt,topsep=4pt,parsep=0pt]
    \item \textbf{\hyperref[sec:appendix_in_the_wild_pose_annotation]{\ref*{sec:appendix_in_the_wild_pose_annotation} Additional Reconstruction Results.}}
    In-the-wild and long-sequence reconstructions, comparisons with ViPE, and runtime analysis.
    \item \textbf{\hyperref[sec:appendix_more_evaluations]{\ref*{sec:appendix_more_evaluations} Additional Trajectory Evaluations.}}
    Results on additional benchmarks, including optional global refinement.
    \item \textbf{\hyperref[sec:appendix_per_sequence_ablations]{\ref*{sec:appendix_per_sequence_ablations} Per-Sequence Results.}}
    Per-sequence comparison results, Per-squence component ablations.
    \item \textbf{\hyperref[sec:appendix_calibration]{\ref*{sec:appendix_calibration} RGB-Only Camera Calibration.}}
    Intrinsic initialization, shared focal calibration, and prefix replay.
    \item \textbf{\hyperref[sec:appendix_focal_refinement]{\ref*{sec:appendix_focal_refinement} Focal Refinement from Revisit Geometry.}}
    Optional revisit-based focal refinement and its evaluation on VBR.
    \item \textbf{\hyperref[sec:implementation]{\ref*{sec:implementation} Implementation Details.}}
    Training and inference settings, feature alignment, loop closure, and stereo integration.
\end{itemize}

\subsection{Additional Reconstruction Results}
\label{sec:appendix_in_the_wild_pose_annotation}

\begin{figure}
    \centering
    \includegraphics[width=\textwidth]{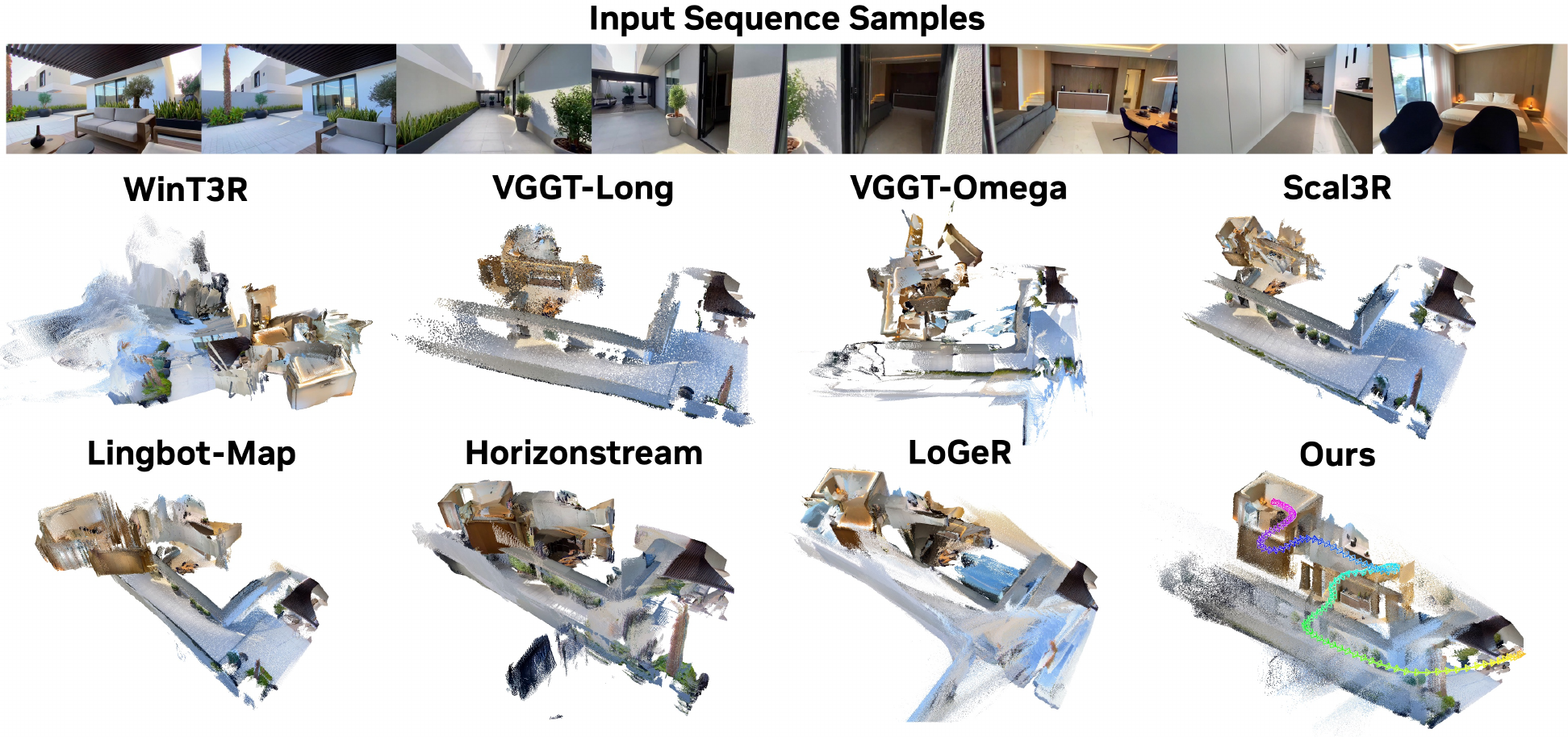}
    \caption{\textbf{In-the-wild reconstruction across scale
    changes.} Representative inputs span an indoor--outdoor--indoor
    transition. The complete reconstructions show whether the indoor and
    outdoor regions remain connected in a common scene.}
    \label{fig:itw_long_sequence}
\end{figure}

\paragraph{Reconstruction across scale changes.}
\cref{fig:itw_long_sequence} shows a handheld sequence that traverses
indoor and outdoor spaces before returning indoors. The resulting changes in
depth range and scene extent test whether local estimates can remain in a
common scale and coordinate frame. Both offline/chunk-based~\citep{scal3r,wang2026vggtomega,vggtlong,li2026wintr} and streaming baselines~\citep{chen2026lingbotmap,cheng2026horizonstream,loger} 
exhibit fragmentation, duplicated structure, or scale-dependent
misalignment. \ourmethod{} instead maintains a coherent scene and camera trajectory
throughout the sequence, demonstrating stable geometric composition beyond a
fixed network context.

\begin{figure}
    \centering
    \includegraphics[width=\textwidth]{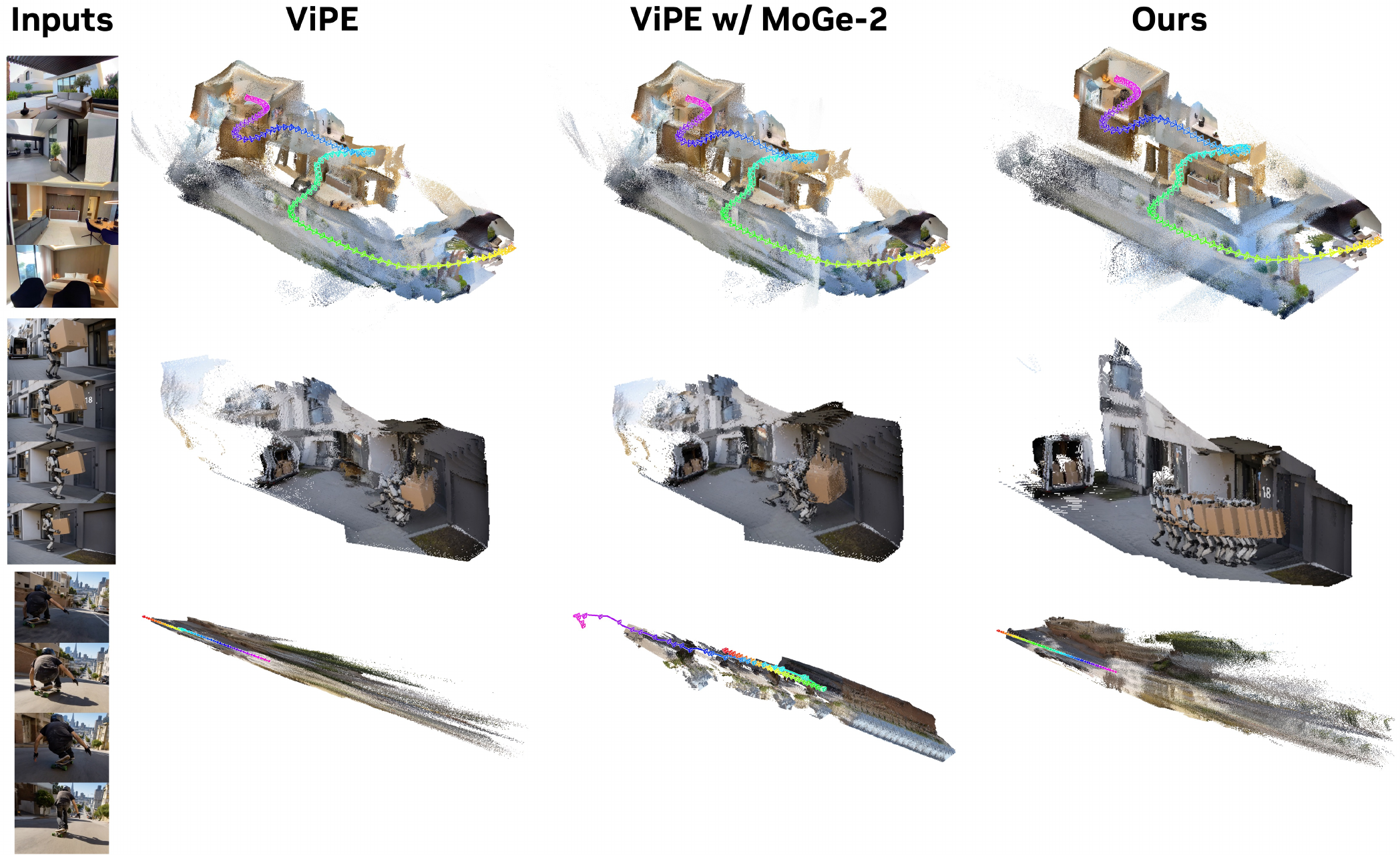}
    \caption{\textbf{Comparison with ViPE on in-the-wild sequences.} 
    Columns show the inputs, ViPE, ViPE with its metric
    prior replaced by MoGe-2, and \ourmethod{}. From top to bottom, the sequences
    contain a pronounced scale change, backward motion with a turn, and rapid
    forward motion. The shared-prior comparison tests whether replacing
    ViPE's depth provider resolves the deformation.}
    \label{fig:itw_vipe}
\end{figure}

\paragraph{In-the-wild comparison with ViPE.}
ViPE may exhaust GPU memory on long sequences, whereas \ourmethod{} processes them
with bounded active memory. We therefore complement the long-sequence results
with a comparison on three shorter but challenging videos for
which both systems can run end to end. 
To separate the contribution of the metric prior from that of the sequential
estimator, we additionally replace ViPE's original prior with the same MoGe-2
prior used by \ourmethod{} while leaving the remainder of its pipeline unchanged
\citep{huang2025vipe,wang2025moge2}. \cref{fig:itw_vipe} considers a
strong scale transition, backward motion followed by a turn, and rapid forward
motion. The two ViPE variants retain visible scene deformation and near--far
misalignment even after sharing the metric prior. \ourmethod{} consistently preserves
sharper structural boundaries and better aligned foreground and background
geometry with substantially less drift. This comparison shows that reliable
reconstruction depends not only on the feed-forward prior, but also on how its
geometry is composed and refined over time.

\begin{figure}
    \centering
    \includegraphics[width=0.9\textwidth]{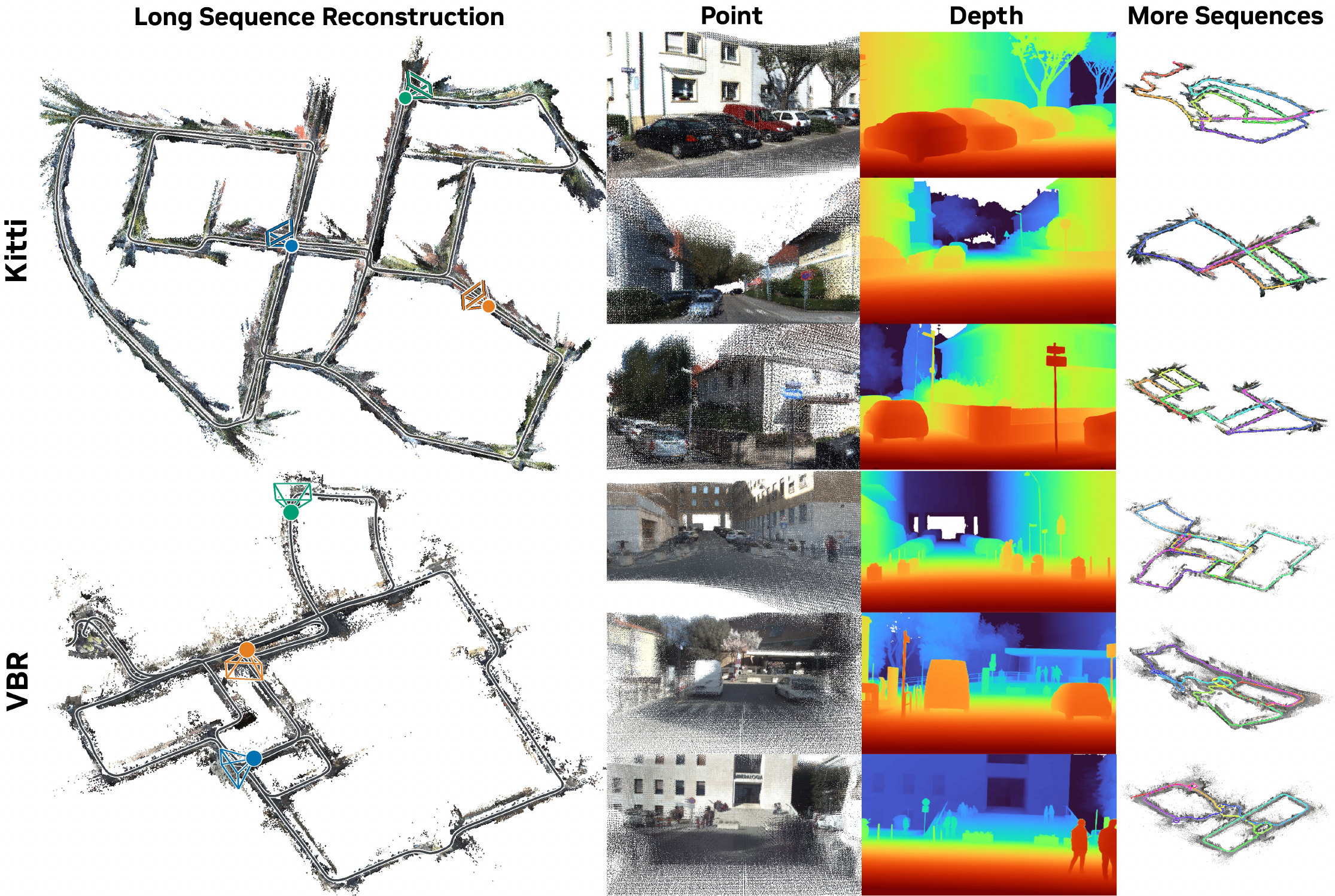}
    \caption{\textbf{Additional long-sequence reconstructions.} KITTI
    examples appear above and VBR examples below. From left to right,
    we show a full-sequence reconstruction with selected camera locations,
    local point clouds, corresponding depth predictions, and
    additional reconstructions.}
    \label{fig:additional_reconstructions}
\end{figure}

\paragraph{Additional long-sequence reconstructions.}
\cref{fig:additional_reconstructions} presents further qualitative
results on KITTI and VBR. The full-sequence views show road networks
reconstructed over extended trajectories, while the selected local views
retain buildings, parked vehicles, and street furniture. Reconstructions
of additional sequences show this behavior across different route layouts.
Together, the global and local views illustrate how predicted metric depth
and geometrically estimated camera motion contribute to reconstruction
at both scene and object scales.

\begin{table}[t]
\centering
\captionsetup{font=small,justification=raggedright,
  singlelinecheck=false}
\caption{\textbf{Runtime on \textbf{KITTI-02} using an NVIDIA A100 GPU.}
Latency is time to first output; FPS is steady-state throughput.}
\label{tab:runtime_analysis}
\footnotesize
\setlength{\tabcolsep}{2.2pt}
\renewcommand{\arraystretch}{1.02}
\resizebox{0.72\linewidth}{!}{%
\begin{tabular}{@{}lcccc@{}}
\toprule
\textbf{Method} &
\makecell{\textbf{ATE(m)} $\downarrow$} &
\textbf{Latency} $\downarrow$ &
\textbf{FPS} $\uparrow$ &
\makecell{\textbf{Peak VRAM (GB)} $\downarrow$} \\
\midrule
LingBot-MAP & 82.29 & \movbestcell{~0.2s} & \movsecondcell{11.37} & 15.07 \\
HorizonStream & 84.62 & ~1.2s & 9.92 & 25.08 \\
Ours (Uncalibrated) & \movsecondcell{24.69} & ~175s & \movbestcell{12.70} & \movsecondcell{6.81} \\
Ours (Calibrated) & \movbestcell{18.83} & \movsecondcell{~0.7s}  & \movbestcell{12.70} & \movbestcell{3.21} \\
\bottomrule
\end{tabular}}
\end{table}

\paragraph{Runtime analysis.}
\cref{tab:runtime_analysis} reports time to first output, steady-state
throughput, and peak allocated GPU memory on KITTI-02 using an NVIDIA A100 GPU.
With supplied intrinsics, \ourmethod{} starts after approximately 0.7 s,
runs at 12.70 FPS, and has a peak GPU memory footprint of 3.21 GB. Estimating the camera model
from RGB introduces a one-time start-up stage, increasing latency to
approximately 175 s and peak memory to 6.81 GB, while subsequent throughput
remains 12.70 FPS. This start-up latency includes the initial MoGe-2 prediction,
geometric camera calibration, and replaying the prefix through the local
estimator.

\subsection{Additional Trajectory Evaluations}
\label{sec:appendix_more_evaluations}

\paragraph{7Scenes camera trajectory comparison.}
Following the LingBot-MAP protocol~\citep{chen2026lingbotmap}, we evaluate all
18 official test sequences of 7Scenes at stride 5,
using RGB frames processed at $434\!\times\!574$ resolution. We report ATE RMSE in
centimeters after a proper per-sequence $\mathrm{Sim}(3)$ alignment and take
the unweighted mean of the 18 sequence-level errors. \cref{tab:seven_scenes_pose}
compares our method with LoGeR, LingBot-MAP, HorizonStream, and
ViPE~\citep{loger,chen2026lingbotmap,cheng2026horizonstream,huang2025vipe}.
Without provided intrinsics, the local \ourmethod{} estimator obtains 6.7 cm.
Optional global bundle adjustment reduces ATE to 6.3 cm, outperforming all
listed baselines except LoGeR$^*$, which introduces further post-hoc alignment.

\begin{table*}[t]
\centering
\caption{\textbf{Full-sequence camera trajectory estimation on \textbf{7Scenes}.} We
report mean ATE (cm; lower is better) over the 18 official test sequences,
sampled at stride 5, after a proper per-sequence $\mathrm{Sim}(3)$ alignment.
Global BA denotes global bundle adjustment.}
\label{tab:seven_scenes_pose}
\small
\setlength{\tabcolsep}{5.0pt}
\renewcommand{\arraystretch}{1.05}
\resizebox{\textwidth}{!}{%
\begin{tabular}{l*{7}{c}}
\toprule
\textbf{Metric} &
\textbf{LoGeR} &
\textbf{LoGeR$^*$} &
\textbf{LingBot-MAP} &
\textbf{HorizonStream} &
\makecell{\textbf{ViPE}\\\textbf{(w/ Global BA)}} &
\makecell{\textbf{Ours}\\\textbf{(Uncalibrated)}} &
\makecell{\textbf{Ours (Uncalibrated)}\\\textbf{w/ Global BA}} \\
\midrule
ATE (cm) $\downarrow$ &
8.1 & \movbestcell{6.1} & 7.9 & 7.1 & 6.5 & 6.7 &
\movsecondcell{6.3} \\
\bottomrule
\end{tabular}}
\end{table*}

\paragraph{Metric global bundle adjustment.}
For the 7Scenes Global BA variant, we preserve the RGB-estimated shared camera
model and the same metric local estimator. Every 15 retained keyframes, the estimator
proposes long-range patch factors from predicted-flow proximity and retains factors that have left
the active window. Whenever this search inserts long-range factors, \ourmethod{} invokes the global BA
solver once. Within this single invocation, the solver performs two optimization
iterations over camera poses and patch depths using the union of active and
retained inactive factors. Poses preceding the earliest active
source remain fixed as gauge anchors. Crucially, \textbf{the global update preserves
the robust metric inverse-depth prior instead of normalizing away the recovered
scale}. Enabling Global BA increases
the measured inference time by only 1.6 seconds per sequence on average over
the 18 test sequences of 7Scenes.

\paragraph{TUM RGB-D camera trajectory comparison.}
We next evaluate all nine Freiburg1 sequences of TUM RGB-D
\citep{sturm2012benchmark}, retaining every associated frame at stride 1 for a
total of 8,694 frames. We report full-trajectory ATE RMSE in centimeters after
a proper per-sequence $\mathrm{Sim}(3)$ alignment and average the nine errors
equally. We compare with DPVO, CUT3R, MASt3R-SLAM, VGGT-SLAM, VGGT-Long,
LoGeR, LingBot-MAP, HorizonStream, and ViPE
\citep{teed2023dpvo,cut3r,murai2025mast3rslam,vggtslam,vggtlong,loger,
chen2026lingbotmap,cheng2026horizonstream,huang2025vipe}.

TUM is particularly challenging for the local DPVO backbone. Unlike the
7Scenes protocol, its input provides \textbf{little translational
baseline, while rapid rotations, rolling shutter, and motion blur create large
but unreliable image motion}. This combination weakens both triangulation and
patch correspondence. By contrast, ViPE inherits DROID-SLAM's dense-flow formulation and
aggregates correspondence evidence over the full image rather than a sparse
set of patches. This broader spatial support makes its pose updates more
resilient when low parallax, rotation, or blur degrades part of the matching
field~\citep{teed2021droid,huang2025vipe}.

\begin{table*}[t]
\centering
\caption{\textbf{Full-sequence camera trajectory estimation on \textbf{TUM RGB-D}.} We report ATE RMSE (cm; lower is better) over nine sequences and 8,694 frames processed at stride 1. Each sequence is aligned independently by
a proper $\mathrm{Sim}(3)$ transform, and sequence-level errors are averaged
equally. DPVO uses supplied intrinsics. Best and
second-best results across both blocks are highlighted.}
\label{tab:tum_rgbd_pose}
\footnotesize
\setlength{\tabcolsep}{2.6pt}
\renewcommand{\arraystretch}{1.05}

\textbf{Prior methods}\par
\resizebox{\textwidth}{!}{%
\begin{tabular}{l*{10}{c}}
\toprule
\textbf{Metric} &
\makecell{\textbf{DPVO}\\\textbf{(Calibrated)}} &
\textbf{CUT3R} &
\makecell{\textbf{MASt3R-}\\\textbf{SLAM}} &
\makecell{\textbf{VGGT-}\\\textbf{SLAM}} &
\makecell{\textbf{VGGT-}\\\textbf{Long}} &
\textbf{LoGeR} &
\textbf{LoGeR$^*$} &
\textbf{LingBot-MAP} &
\textbf{HorizonStream} &
\makecell{\textbf{ViPE}\\\textbf{(Uncalibrated)}} \\
\midrule
ATE (cm) $\downarrow$ &
9.7 & 48.6 & 6.0 & 6.1 & 8.3 & 7.5 & 6.0 &
\movsecondcell{\underline{4.5}} & 5.3 & 4.9 \\
\bottomrule
\end{tabular}}

\textbf{\ourmethod{} variants}\par
\resizebox{0.85\linewidth}{!}{%
\begin{tabular}{l*{6}{c}}
\toprule
& \multicolumn{3}{c}{\textbf{Uncalibrated}} &
\multicolumn{3}{c}{\textbf{Calibrated}} \\
\cmidrule(lr){2-4}\cmidrule(lr){5-7}
\textbf{Metric} & \textbf{Ours} & \textbf{w/ Global BA} &
\textbf{w/ Global BA, Refine} & \textbf{Ours} & \textbf{w/ Global BA} &
\textbf{w/ Global BA, Refine} \\
\midrule
ATE (cm) $\downarrow$ & 8.4 & 4.9 & \movbestcell{\underline{4.4}} &
7.5 & \movsecondcell{4.1} & \movbestcell{3.7} \\
\bottomrule
\end{tabular}}
\end{table*}

\paragraph{Global refinement on TUM RGB-D.}
For dense TUM streams, we optimize active and retained historical patch
factors every eight keyframes, propagate revised keyframe poses to culled
frames, and remeasure stored targets and confidences before two terminal
solves. Camera poses and patch depths are refined jointly while preserving the
metric-depth prior and sequence scale. Our \emph{w/ Refine} variant further
introduces a temporally smooth, zero-mean log-scale correction to each
keyframe's depth prior, bidirectional long-range factors with
connectivity-aware culling, and recurrent-state warm-starting during
remeasurement. The base terminal passes add about 3.6 seconds per sequence.
Relative to Global BA, \emph{w/ Refine} increases the mean geometric-backend
time by 5.3 seconds while leaving the local estimator and camera model
unchanged. As shown in \cref{tab:tum_rgbd_pose}, our local estimator improves the
calibrated DPVO backbone from 9.7 cm to 7.5 cm. Global BA reduces the error to
4.1 cm, and \emph{w/ Refine} reaches 3.7 cm---a 62\% reduction over DPVO.
Without provided intrinsics, the same progression is 8.4 cm, 4.9 cm, and
4.4 cm, respectively, with the refined variant outperforming all RGB-only
methods in the comparison. This result shows that revisiting both historical
states and their measurements converts accurate local estimates into a
globally consistent trajectory.

\paragraph{LaMAria camera trajectory comparison.}
We evaluate on five full-length sequences from
LaMAria~\citep{krishnan2025lamaria}: \texttt{R\_01}, \texttt{R\_04},
\texttt{R\_06}, \texttt{R\_07}, and \texttt{R\_08}, which span easy,
medium, and hard trajectories. Unlike the RGB benchmarks in the main paper,
LaMAria is evaluated from monocular grayscale input, creating a pronounced
appearance-domain shift. We use no LaMAria-specific training or fine-tuning.
We follow the same full-sequence protocol, reporting ATE RMSE in meters after
a proper $\mathrm{Sim}(3)$ alignment and averaging the five sequence-level
errors equally.

\begin{table}[!htbp]
\centering
\caption{\textbf{Full-sequence ATE (m) on five sequences from
\textbf{LaMAria}.}}
\label{tab:lamaria}
\scriptsize
\setlength{\tabcolsep}{3.0pt}
\resizebox{0.8\linewidth}{!}{%
\begin{tabular}{l*{5}{c}c}
\toprule
\textbf{Method} &
\makecell{\textbf{R\_01}\\2,898 fr.} &
\makecell{\textbf{R\_04}\\5,253 fr.} &
\makecell{\textbf{R\_06}\\7,808 fr.} &
\makecell{\textbf{R\_07}\\8,010 fr.} &
\makecell{\textbf{R\_08}\\12,328 fr.} &
\textbf{Avg.} $\downarrow$ \\
\midrule
LingBot-MAP & 6.65 & 33.01 & 60.85 & 62.36 & 98.91 & 52.36 \\
LingBot-MAP-Long & 7.35 & 34.82 & 67.10 & 65.37 & 77.24 & 50.38 \\
HorizonStream & 7.19 & 35.63 & 54.55 & 42.70 & 90.63 & 46.14 \\
LoGeR & 4.23 & 10.87 & 22.44 & 23.01 & 91.11 & 30.33 \\
LoGeR$^*$ & 4.63 & 14.27 & 18.76 & 35.41 & 69.81 & 28.58 \\
\midrule\midrule
Ours (Uncalibrated) & \movsecondcell{3.09} & \movsecondcell{4.85} & \movsecondcell{15.97} & \movbestcell{10.14} & \movsecondcell{9.57} & \movsecondcell{8.73} \\
Ours (Calibrated) & \movbestcell{2.69} & \movbestcell{4.39} & \movbestcell{15.03} & \movsecondcell{11.50} & \movbestcell{9.30} & \movbestcell{8.58} \\
\bottomrule
\end{tabular}}
\end{table}

\ourmethod{} achieves average ATEs of 8.73 m and 8.58 m in the uncalibrated and
calibrated settings, respectively, compared with 28.58 m for the strongest
baseline.
It obtains the best result on every sequence under at least one calibration
setting. The large margin under grayscale input suggests that systems whose
long-horizon state is formed entirely by learned prediction can be more
sensitive to appearance-domain shift. In contrast, \ourmethod{} combines transferable
feed-forward evidence with explicit optimization of the sequence-specific
constraints. Without LaMAria adaptation, these results show
out-of-domain generalization rather than benchmark-specific training.

\paragraph{Wild-SLAM MoCap camera trajectory comparison.}
We evaluate trajectories on all ten dynamic Wild-SLAM MoCap
sequences~\citep{zheng2025wildgsslam}. Following the WildPose
protocol~\citep{zheng2026wildpose}, we process every associated RGB frame
(stride 1; 9,039 frames in total), evaluate the complete per-frame trajectory,
and report translational ATE RMSE in centimeters after a proper per-sequence
$\mathrm{Sim}(3)$ Umeyama alignment. The ten sequence-level errors are
averaged equally. We compare with LingBot-MAP,
ViPE, HorizonStream,
DROID-SLAM, Splat-SLAM~\citep{sandstrom2025splatslam},
and WildPose.

Wild-SLAM MoCap contains substantial \textbf{independently moving content and
occlusion}, making it particularly favorable to estimators designed to suppress
dynamic observations. The evaluated ViPE configuration filters predefined
movable semantics with instance masks, while WildPose predicts learned motion
masks. In contrast, the reported \ourmethod{} variants use the same estimator as on
the static benchmarks, without an explicit dynamic-object mask or a
dynamics-specific objective. Nevertheless, \ourmethod{} reaches 2.70 cm with
RGB-only calibration, outperforming LingBot-MAP and HorizonStream and
approaching the dynamic-aware ViPE result without a dynamics-specific
component. With supplied intrinsics, its error decreases to 1.65 cm,
outperforming DROID-SLAM and Splat-SLAM; only WildPose, with dedicated learned
motion masking, remains ahead. These results demonstrate transfer to dynamic
video without task-specific modifications.

\begin{table*}[t]
\centering
\caption{\textbf{Full-trajectory camera pose estimation on the dynamic
\textbf{Wild-SLAM MoCap} benchmark.} We report ATE RMSE (cm; lower is better),
averaged equally over ten sequences. Methods are separated by whether camera
intrinsics are provided at inference. $\dagger$ marks evaluated configurations
that explicitly mask dynamic content. Best and second-best results are
highlighted separately within each calibration regime. Our two variants use
neither dynamic masks nor a dynamics-specific objective and operate without
loop closure or global bundle adjustment.}
\label{tab:wild_slam_mocap}
\small
\setlength{\tabcolsep}{4.0pt}
\renewcommand{\arraystretch}{1.05}
\resizebox{\textwidth}{!}{%
\begin{tabular}{l*{8}{c}}
\toprule
& \multicolumn{4}{c}{\textbf{Uncalibrated}} &
\multicolumn{4}{c}{\textbf{Calibrated}} \\
\cmidrule(lr){2-5}\cmidrule(lr){6-9}
\textbf{Metric} &
\textbf{LingBot-MAP} &
\textbf{ViPE$^\dagger$} &
\textbf{HorizonStream} &
\textbf{Ours} &
\textbf{DROID-SLAM} &
\textbf{Splat-SLAM} &
\textbf{WildPose$^\dagger$} &
\makecell{\textbf{Ours}} \\
\midrule
ATE (cm) $\downarrow$ &
3.48 & \movbestcell{2.59} & 21.58 & \movsecondcell{2.70} &
16.17 & 8.71 & \movbestcell{0.39} &
\movsecondcell{1.65} \\
\bottomrule
\end{tabular}}
\end{table*}

\subsection{Per-Sequence Results}
\label{sec:appendix_per_sequence_ablations}

\paragraph{Per-Sequence Comparison of KITTI, Oxford Spires, and VBR.}
\cref{tab:kitti_odometry,tab:oxford_spires,tab:vbr}
report full-sequence ATE on KITTI Odometry, Oxford Spires, and VBR,
respectively, providing the per-sequence breakdown of the benchmark averages
in the main paper. Baseline sources are VGGT-Long~\citep{vggtlong} for
$\pi^3$-Long and DA3-Long, and VGGT-SLAM~\citep{vggtslam} for DA3-SLAM.
\begin{table*}[t]
\centering
\caption{\textbf{Full-sequence ATE (m) on KITTI Odometry.} $\dagger$ marks method-native sparse output over keyframes. LoGeR$^*$ requires alignment
optimization. The strongest baseline average within each method group and our default RGB-only configuration are underlined.}
\label{tab:kitti_odometry}
\scriptsize
\setlength{\tabcolsep}{2.1pt}
\resizebox{\textwidth}{!}{%
\begin{tabular}{cl*{11}{c}c}
\toprule
& \textbf{Method} &
\makecell{\textbf{00}\\4,541 fr.\\3.7 km} &
\makecell{\textbf{01}\\1,101 fr.\\2.5 km} &
\makecell{\textbf{02}\\4,661 fr.\\5.1 km} &
\makecell{\textbf{03}\\801 fr.\\0.6 km} &
\makecell{\textbf{04}\\271 fr.\\0.4 km} &
\makecell{\textbf{05}\\2,761 fr.\\2.2 km} &
\makecell{\textbf{06}\\1,101 fr.\\1.2 km} &
\makecell{\textbf{07}\\1,101 fr.\\0.7 km} &
\makecell{\textbf{08}\\4,071 fr.\\3.2 km} &
\makecell{\textbf{09}\\1,591 fr.\\1.7 km} &
\makecell{\textbf{10}\\1,201 fr.\\0.9 km} & \textbf{Avg.} $\downarrow$ \\
\midrule
\multirow{8}{*}{\rotatebox{90}{Optimization}}
& VidMap$\dagger$ (Uncalibrated) & 6.89 & 424.80 & 169.49 & 4.74 & 25.79 & 80.18 & 132.81 & 77.72 & 120.01 & 176.84 & 85.95 & 118.66 \\
& DROID-SLAM (Calibrated) & 131.25 & 267.79 & 217.63 & 3.40 & 1.47 & 80.41 & 64.20 & 18.19 & 167.80 & 90.05 & 22.74 & 96.81 \\
& DPVO (Calibrated) & 113.11 & 16.60 & 113.01 & 2.46 & 0.98 & 59.34 & 55.91 & 19.30 & 110.63 & 74.55 & 13.71 & 52.69 \\
& VidMap$\dagger$ (Calibrated) & 7.50 & 234.88 & 35.74 & 2.00 & 14.66 & 1.66 & 49.00 & 13.50 & 123.27 & 8.97 & 2.87 & 44.91 \\
& DROID-W (Calibrated) & 48.26 & 207.57 & 52.80 & 1.76 & 4.65 & 25.16 & 5.34 & 3.35 & 29.05 & 35.56 & 8.01 & 38.32 \\
& COLMAP & 139.12 & \movbestcell{3.83} & 71.99 & \movsecondcell{1.46} & 112.77 & 20.37 & 10.95 & 7.80 & 21.72 & 21.19 & 4.52 & 37.79 \\
& ViPE (Uncalibrated) & 25.87 & 148.34 & 53.35 & 5.29 & 1.95 & 24.22 & 11.72 & \movsecondcell{1.03} & 18.51 & 20.55 & 4.71 & 28.68 \\
& ViPE (Calibrated) & 24.31 & 25.42 & 29.15 & 3.94 & 5.87 & 10.37 & 27.34 & 5.38 & 25.34 & 36.24 & 6.10 & \underline{18.13} \\
\midrule
\multirow{8}{*}{\rotatebox{90}{Offline/Chunk}}
& MASt3R-SLAM & 190.12 & 507.15 & 232.17 & 169.82 & 88.98 & 126.37 & 133.33 & 62.92 & 199.70 & 178.03 & 151.69 & 185.48 \\
& VGGT-SLAM & 27.99 & 214.23 & 260.88 & 53.50 & 26.02 & 99.61 & 23.62 & 64.52 & 243.87 & 169.10 & 80.13 & 114.86 \\
& VGGT-Long & 8.64 & 61.21 & 52.72 & 8.78 & 4.20 & 9.88 & 4.67 & 2.66 & 72.98 & 31.84 & 27.71 & 25.94 \\
& LoGeR & 54.98 & 36.57 & 36.20 & 4.27 & 1.62 & 33.41 & 11.78 & 13.33 & 22.92 & 17.89 & 8.06 & 21.91 \\
& $\pi^3$-Long & 5.55 & 114.83 & 50.29 & 1.63 & 1.11 & 3.48 & 2.88 & 3.92 & 24.25 & 7.38 & 17.61 & 21.17 \\
& DA3-Long & 5.13 & 78.76 & 35.64 & 5.38 & 3.18 & 3.04 & 2.83 & 2.32 & 26.55 & 8.86 & 13.42 & 16.83 \\
& LoGeR$^*$ & 26.19 & 41.26 & 32.21 & 5.02 & 1.62 & 22.65 & 5.49 & 5.04 & 21.96 & 9.03 & 9.44 & 16.35 \\
& Scal3R w/ LC & 4.30 & 45.29 & 42.06 & 3.36 & 1.74 & 3.30 & 2.49 & 2.03 & 36.69 & 12.32 & 6.46 & \underline{14.55} \\
\midrule
\multirow{7}{*}{\rotatebox{90}{Online}}
& STream3R & 190.98 & 681.95 & 301.40 & 158.25 & 102.73 & 159.85 & 135.03 & 90.37 & 261.15 & 216.31 & 207.49 & 227.77 \\
& StreamVGGT & 191.93 & 653.06 & 303.35 & 157.50 & 108.24 & 160.46 & 133.71 & 89.00 & 263.95 & 216.69 & 209.80 & 226.15 \\
& InfiniteVGGT & 167.17 & 533.36 & 272.99 & 149.18 & 58.86 & 127.50 & 100.54 & 78.77 & 196.66 & 199.25 & 138.04 & 183.85 \\
& LongStream & 92.55 & 46.01 & 134.70 & 3.81 & 1.95 & 84.69 & 23.12 & 14.93 & 62.07 & 85.61 & 21.48 & 51.90 \\
& LingBot-MAP & 30.80 & 64.74 & 82.29 & 2.49 & 0.85 & 16.55 & 6.27 & 8.92 & 39.32 & 17.99 & 7.96 & 25.29 \\
& HorizonStream & 26.40 & 20.62 & 84.62 & 5.15 & 0.62 & 12.82 & 4.59 & 5.49 & 19.49 & 25.73 & 11.71 & 19.75 \\
& HorizonStream w/ LC & 13.91 & 20.62 & 69.43 & 5.15 & 0.62 & 6.86 & 6.50 & 2.67 & 19.49 & 23.86 & 11.71 & \underline{16.44} \\
\midrule\midrule
\multirow{6}{*}{\rotatebox{90}{Ours}}
& Ours (Uncalibrated) & 15.29 & 35.94 & 24.69 & 1.48 & 0.69 & 18.43 & 9.14 & 3.81 & 22.32 & 14.50 & 8.21 & \underline{\textbf{14.05}} \\
& Ours (Uncalibrated) w/ LC & \movsecondcell{2.75} & 35.94 & 18.36 & 1.48 & 0.69 & 4.44 & 5.61 & 2.10 & 22.32 & 5.64 & 8.21 & 9.78 \\
& Ours (Calibrated) & 5.22 & 33.21 & 18.83 & 2.71 & \movsecondcell{0.59} & 5.90 & 6.18 & 3.13 & \movsecondcell{11.94} & 8.39 & \movsecondcell{3.95} & 9.10 \\
& Ours (Calibrated) w/ LC & 2.92 & 33.21 & \movbestcell{6.02} & 2.71 & \movsecondcell{0.59} & 3.32 & 9.49 & 2.48 & \movsecondcell{11.94} & \movsecondcell{4.69} & \movsecondcell{3.95} & 7.39 \\
& Ours (Stereo) w/o LC & 5.21 & \movsecondcell{6.33} & \movsecondcell{12.62} & \movbestcell{0.34} & \movbestcell{0.23} & \movsecondcell{2.76} & \movsecondcell{1.45} & \movbestcell{0.97} & \movbestcell{5.27} & \movbestcell{4.14} & \movbestcell{0.93} & \movsecondcell{3.66} \\
& Ours (Stereo) w/ LC & \movbestcell{1.43} & \movsecondcell{6.33} & \movsecondcell{12.62} & \movbestcell{0.34} & \movbestcell{0.23} & \movbestcell{0.55} & \movbestcell{0.20} & \movbestcell{0.97} & \movbestcell{5.27} & \movbestcell{4.14} & \movbestcell{0.93} & \movbestcell{3.00} \\
\bottomrule
\end{tabular}}
\end{table*}

\begin{table*}[t]
\centering
\caption{\textbf{Full-sequence ATE (m) on Oxford Spires at stride one.}}
\label{tab:oxford_spires}
\scriptsize
\setlength{\tabcolsep}{3.0pt}
\resizebox{\textwidth}{!}{%
\begin{tabular}{cl*{10}{c}c}
\toprule
& \textbf{Method} &
\makecell{\textbf{Keble-2}\\0.20 km} &
\makecell{\textbf{Keble-3}\\0.18 km} &
\makecell{\textbf{Keble-4}\\0.23 km} &
\makecell{\textbf{Keble-5}\\0.21 km} &
\makecell{\textbf{Observ.-1}\\0.27 km} &
\makecell{\textbf{Observ.-2}\\0.28 km} &
\makecell{\textbf{Christ-2}\\0.25 km} &
\makecell{\textbf{Christ-3}\\0.18 km} &
\makecell{\textbf{Christ-5}\\0.80 km} &
\makecell{\textbf{Bodleian-2}\\0.55 km} &
\textbf{Avg.} $\downarrow$ \\
\midrule
\multirow{7}{*}{\rotatebox{90}{Offline/Chunk}}
& VGGT-SLAM & 20.56 & 16.20 & 24.71 & 30.02 & 3.97 & 7.42 & 30.09 & 4.90 & 28.13 & 68.24 & 23.42 \\
& MASt3R-SLAM & 7.06 & 18.03 & 11.27 & 14.64 & 10.26 & 10.30 & 2.90 & 6.95 & 19.85 & 19.99 & 12.12 \\
& DA3-SLAM & 2.86 & 3.88 & 4.34 & 7.97 & 18.69 & 3.87 & 8.89 & 6.93 & 15.48 & 44.78 & 11.77 \\
& VGGT-Long & 9.07 & 9.37 & 11.77 & 3.96 & 6.03 & 4.05 & \movsecondcell{2.41} & \movsecondcell{0.48} & 14.38 & 12.59 & 7.41 \\
& ViPE & 14.39 & 13.88 & \movbestcell{0.57} & 12.18 & \movbestcell{0.55} & \movbestcell{0.23} & 15.44 & \movbestcell{0.18} & \movsecondcell{9.31} & \movsecondcell{3.89} & 7.06 \\
& LoGeR & 2.74 & 2.33 & 2.92 & 3.21 & 4.83 & 5.29 & \movbestcell{2.14} & 1.84 & 11.61 & 6.31 & 4.32 \\
& LoGeR$^*$ & \movsecondcell{2.12} & \movbestcell{1.85} & 1.70 & \movsecondcell{1.61} & 6.05 & 7.31 & 2.48 & 0.98 & 14.38 & 4.31 & \movsecondcell{4.28} \\
\midrule
\multirow{7}{*}{\rotatebox{90}{Online}}
& Stream3R-w & 31.28 & 22.39 & 31.07 & 24.63 & 28.38 & 27.18 & 38.12 & 12.99 & 42.39 & 82.66 & 34.11 \\
& InfiniteVGGT & 25.02 & 20.39 & 25.09 & 24.15 & 29.10 & 28.14 & 33.02 & 12.95 & 40.39 & 79.32 & 31.76 \\
& LongStream & 11.57 & 9.44 & 16.88 & 10.67 & 12.43 & 14.34 & 17.93 & 6.42 & 35.49 & 24.02 & 15.92 \\
& CUT3R & 13.10 & 11.55 & 15.76 & 8.98 & 13.58 & 15.60 & 12.36 & 5.71 & 31.31 & 19.42 & 14.74 \\
& TTT3R & 8.27 & 7.26 & 8.77 & 7.55 & 9.43 & 10.30 & 4.56 & 2.11 & 21.75 & 14.00 & 9.40 \\
& HorizonStream & 5.99 & 6.17 & 7.52 & 6.23 & 5.89 & 5.26 & 10.64 & 3.03 & 25.48 & 11.25 & 8.75 \\
& LingBot-MAP & 2.66 & 2.25 & 1.56 & 2.45 & 3.23 & 2.61 & 2.89 & 0.70 & 27.27 & 5.87 & 5.15 \\
\midrule\midrule
& Ours (Uncalibrated)
& \movbestcell{0.53}
& \movsecondcell{1.88}
& \movsecondcell{1.39}
& \movbestcell{0.49}
& \movsecondcell{0.76}
& \movsecondcell{1.04}
& \movbestcell{2.14}
& 1.19
& \movbestcell{5.71}
& \movbestcell{2.53}
& \movbestcell{1.77} \\
\bottomrule
\end{tabular}}
\end{table*}

\begin{table*}[t]
\centering
\caption{\textbf{Full-sequence ATE (m) on \textbf{VBR}.} The header reports the number
of frames and route length of each sequence.}
\label{tab:vbr}
\scriptsize
\setlength{\tabcolsep}{4.0pt}
\resizebox{0.98\textwidth}{!}{%
\begin{tabular}{cl*{7}{c}c}
\toprule
& \textbf{Method} &
\makecell{\textbf{colosseo\_0}\\8,815 fr.\\1.45 km} &
\makecell{\textbf{campus\_0}\\12,042 fr.\\2.73 km} &
\makecell{\textbf{campus\_1}\\11,671 fr.\\2.95 km} &
\makecell{\textbf{pincio\_0}\\11,142 fr.\\1.27 km} &
\makecell{\textbf{spagna\_0}\\14,141 fr.\\1.56 km} &
\makecell{\textbf{diag\_0}\\10,021 fr.\\1.02 km} &
\makecell{\textbf{ciampino\_1}\\18,846 fr.\\5.20 km} &
\textbf{Avg.} $\downarrow$ \\
\midrule
\multirow{6}{*}{\rotatebox{90}{Offline/Chunk}}
& MASt3R-SLAM & 88.75 & 116.21 & 114.37 & 67.11 & 54.62 & 35.67 & 171.76 & 92.64 \\
& VGGT-Long & 45.73 & 132.68 & 115.65 & 64.40 & 58.49 & 33.66 & 187.97 & 91.23 \\
& VGGT-SLAM & 101.00 & 93.51 & 71.74 & 66.42 & 57.00 & 33.64 & 124.10 & 78.20 \\
& ViPE  & 79.22 & 46.07 & 63.10 & 15.42 & 20.33 & 35.75 & 63.48 & 46.20 \\
& LoGeR & 31.77 & 27.90 & 30.80 & 17.96 & 21.33 & 32.25 & 34.16 & 28.02 \\
& LoGeR$^*$ & 55.32 & 13.27 & 16.79 & 9.18 & 18.32 & 29.45 & 34.32 & 25.24 \\
\midrule
\multirow{7}{*}{\rotatebox{90}{Online}}
& InfiniteVGGT & 89.89 & 136.76 & 116.63 & 77.76 & 63.35 & 33.84 & 202.09 & 102.90 \\
& LongStream & 72.52 & 100.57 & 105.55 & 43.47 & 59.31 & 32.35 & 131.78 & 77.93 \\
& CUT3R & 82.63 & 42.25 & 43.16 & 46.65 & 44.62 & 28.62 & 175.83 & 66.25 \\
& TTT3R & 75.52 & 59.44 & 56.55 & 33.87 & 37.33 & \movsecondcell{18.49} & 173.71 & 64.99 \\
& LingBot-MAP & \movsecondcell{16.70} & 23.61 & 10.37 & 29.37 & 24.29 & 24.12 & 64.24 & 27.53 \\
& HorizonStream & 37.42 & 22.46 & 22.49 & 22.63 & 23.52 & 22.46 & 26.10 & 25.30 \\
& HorizonStream w/ LC & \movbestcell{12.76} & 28.54 & \movbestcell{8.49} & 17.24 & 23.06 & 24.05 & 17.76 & 18.84 \\
\midrule\midrule
& Ours (Uncalibrated) & 30.08 & \movsecondcell{9.26} & 15.07 & \movsecondcell{7.43} & \movsecondcell{13.94} & 20.97 & \movsecondcell{17.75} & \movsecondcell{16.36} \\
& Ours (Uncalibrated) w/ LC & 26.83 & \movbestcell{6.80} & \movsecondcell{9.99} & \movbestcell{6.94} & \movbestcell{10.18} & \movbestcell{5.84} & \movbestcell{9.60} & \movbestcell{10.88} \\
\bottomrule
\end{tabular}}
\end{table*}

\paragraph{Per-Sequence Component Ablations.}
\cref{tab:component_ablation_per_sequence} reports calibrated monocular
ATE on KITTI Odometry, Oxford Spires, and VBR with the same MoGe-2 prior
and no loop closure. Multi-dataset adaptation broadens the range of camera
motions and optical-flow patterns seen during training, reducing average
ATE from 12.77 to 10.32 m on KITTI and from 1.85 to 1.76 m on Oxford
Spires, but increasing VBR ATE from 19.57 to 22.51 m. Modeling
correspondences across this more heterogeneous training mixture places
greater demands on DPVO's compact frontend. DINOv3 supplies complementary
visual representations from large-scale pretraining, enriching both patch
matching and recurrent context. With this augmentation, average ATE falls
to 9.10, 1.71, and 18.90 m, respectively. The gains across all three
datasets suggest that diverse motion supervision and richer visual
representations work together to improve correspondence generalization.

\begin{table}[!htbp]
\centering
\captionsetup{font=small,justification=raggedright,singlelinecheck=false}
\caption{\textbf{Per-sequence component ablations: calibrated ATE RMSE (m) on
\textbf{KITTI Odometry}, \textbf{Oxford Spires}, and \textbf{VBR}.}
All variants use monocular input, the same MoGe-2 prior, and no loop closure;
Oxford Spires uses stride one. Only the lowest average within each dataset
is bold.}
\label{tab:component_ablation_per_sequence}
\scriptsize
\setlength{\tabcolsep}{2.1pt}
\renewcommand{\arraystretch}{1.02}
\begin{tabular*}{\linewidth}{@{\extracolsep{\fill}}l*{11}{c}c@{}}
\toprule
\multicolumn{13}{@{}l}{\textbf{KITTI Odometry}} \\
\midrule
\textbf{Variants} & \textbf{00} & \textbf{01} & \textbf{02} & \textbf{03} &
\textbf{04} & \textbf{05} & \textbf{06} & \textbf{07} & \textbf{08} &
\textbf{09} & \textbf{10} & \textbf{Avg.} $\downarrow$ \\
\midrule
Ours w/o Multi-dataset, DINO & 7.61 & 76.22 & 17.17 & 2.54 & 0.79 & 8.02 & 5.44 & 2.83 & 10.38 & 4.80 & 4.64 & 12.77 \\
Ours w/o DINO & 6.05 & 30.52 & 21.36 & 2.20 & 3.67 & 4.92 & 10.19 & 3.83 & 17.84 & 7.98 & 5.00 & 10.32 \\
Ours & 5.22 & 33.21 & 18.83 & 2.71 & 0.59 & 5.90 & 6.18 & 3.13 & 11.94 & 8.39 & 3.95 & \textbf{9.10} \\
\bottomrule
\end{tabular*}
\par
\begin{tabular*}{\linewidth}{@{\extracolsep{\fill}}l*{10}{c}c@{}}
\toprule
\multicolumn{12}{@{}l}{\textbf{Oxford Spires}} \\
\midrule
\textbf{Variants} & \textbf{keb.02} & \textbf{keb.03} & \textbf{keb.04} &
\textbf{keb.05} & \textbf{obs.01} & \textbf{obs.02} & \textbf{chr.02} &
\textbf{chr.03} & \textbf{chr.05} & \textbf{bod.02} & \textbf{Avg.} $\downarrow$ \\
\midrule
Ours w/o Multi-dataset, DINO & 0.44 & 2.17 & 0.86 & 0.85 & 0.81 & 0.87 & 2.38 & 1.27 & 5.98 & 2.85 & 1.85 \\
Ours w/o DINO & 0.93 & 1.47 & 0.86 & 0.82 & 1.37 & 0.91 & 1.78 & 1.49 & 5.46 & 2.47 & 1.76 \\
Ours & 0.48 & 1.73 & 0.78 & 0.92 & 0.73 & 0.73 & 2.19 & 1.20 & 5.80 & 2.52 & \textbf{1.71} \\
\bottomrule
\end{tabular*}
\par
\begin{tabular*}{\linewidth}{@{\extracolsep{\fill}}l*{7}{c}c@{}}
\toprule
\multicolumn{9}{@{}l}{\textbf{VBR}} \\
\midrule
\textbf{Variants} & \textbf{colosseo\_0} & \textbf{campus\_0} &
\textbf{campus\_1} & \textbf{pincio\_0} & \textbf{spagna\_0} &
\textbf{diag\_0} & \textbf{ciampino\_1} & \textbf{Avg.} $\downarrow$ \\
\midrule
Ours w/o Multi-dataset, DINO & 19.88 & 16.59 & 15.51 & 7.91 & 15.69 & 21.08 & 40.36 & 19.57 \\
Ours w/o DINO & 10.36 & 22.95 & 22.49 & 6.51 & 18.28 & 24.19 & 52.81 & 22.51 \\
Ours & 14.26 & 17.32 & 15.18 & 6.32 & 13.83 & 22.52 & 42.83 & \textbf{18.90} \\
\bottomrule
\end{tabular*}
\end{table}

\subsection{RGB-Only Camera Calibration}
\label{sec:appendix_calibration}

\paragraph{Feed-forward initialization.}
When intrinsics are unavailable, MoGe-2 estimates a camera matrix
$\widehat{\mathbf{K}}_i$ for each of the first $N_0=32$ RGB frames without
field-of-view conditioning. We initialize
$\mathbf{K}^{(0)}$ with the component-wise median of
$\{\widehat{\mathbf{K}}_i\}_{i=1}^{N_0}$, requiring no ground-truth calibration.

\paragraph{Shared focal calibration.}
The subsequent calibration-only estimator uses 128 keyframes selected from an
initial prefix of $B$ raw frames. The keyframe count is fixed, whereas $B$
depends on keyframe selection in each sequence. Let
$\mathcal{X}_{\mathrm{cal}}$ denote its auxiliary
pose-and-depth state and $\mathcal{E}_{\mathrm{cal}}$ the objective combining
learned-correspondence reprojection residuals with metric-depth constraints.
Writing the entries of $\mathbf{K}^{(0)}$ as
$(f_x^{(0)},f_y^{(0)},c_x^{(0)},c_y^{(0)})$, we optimize
\begin{equation}
\begin{aligned}
    \mathbf{K}(\alpha)
    &=
    \begin{bmatrix}
        f_x^{(0)}+\alpha & 0 & c_x^{(0)}\\
        0 & f_y^{(0)}+\alpha & c_y^{(0)}\\
        0 & 0 & 1
    \end{bmatrix},\\
    (\alpha^{\star},\mathcal{X}_{\mathrm{cal}}^{\star})
    &=
    \arg\min_{\alpha,\,\mathcal{X}_{\mathrm{cal}}}
    \mathcal{E}_{\mathrm{cal}}
    \!\left(
        \mathcal{X}_{\mathrm{cal}},
        \mathbf{K}(\alpha),
        \mathcal{I}_{1:B}
    \right).
    \label{eq:rgb_calibration}
\end{aligned}
\end{equation}
The same scalar $\alpha$ updates both focal axes, while the principal point is
held fixed. Because the metric predictor is conditioned on field of view, its
depth measurements are recomputed whenever the focal estimate changes. The
optimized poses and depths are used only to make the shared focal parameter
observable under multi-view constraints and are not retained as reconstruction
output.

We refer to the feed-forward initialization and shared-focal optimization above
collectively as base calibration. We then discard the auxiliary poses and
depths, fix $\mathbf{K}$, and replay the complete prefix through the regular
local estimator, so that all reported poses come from the same estimator.
Latency is measured as time to first output and includes the initial MoGe-2
camera prediction, DROID-style geometric calibration, and prefix replay.
In the KITTI-02 runtime evaluation, the
uncalibrated system reaches its 6.81 GB peak during calibration; GPU memory
usage drops to 3.21 GB during subsequent steady-state estimation.
This estimate is sufficient when the centered-principal-point model is
consistent with the released imagery. Some sequences, however, require an
additional RGB-only focal refinement, which we describe next.

\subsection{Focal Refinement from Revisit Geometry}
\label{sec:appendix_focal_refinement}

\paragraph{Motivation.}
The base calibration fits a shared focal with the principal point fixed at
the image center. This matches the processed Oxford Spires imagery but
leaves a 17.3-pixel horizontal offset (3.4\% of image width) on VBR.
Under predominantly forward motion, focal and principal-point offsets can
partially compensate in the bearing $((u-c_x)/f,(v-c_y)/f,1)$, allowing
biased depth and translation to produce small reprojection errors.
Retrospective evaluation shows a focal error of 0.1--0.5\% on Oxford
Spires but a 7--9\% overestimate on six of the seven VBR sequences.
We therefore refine the focal using revisits with more varied viewing
directions. The refinement was motivated by our analysis of VBR, 
but can also benefit other sequences whose revisits provide sufficiently diverse 
viewing geometry to constrain the focal length.

\paragraph{RGB-only revisit refinement.}
We use the base trajectory to select temporally distant revisit pairs,
extract RGB correspondences with RoMa~\citep{edstedt2024roma}, and estimate
fundamental matrices $\mathbf{F}_{ij}$ using MAGSAC~\citep{barath2020magsac}.
Let $\mathcal{R}$ be the retained pair set and $\mathbf{K}(f,c_x)$ the
camera matrix with shared focal $f$, horizontal principal point $c_x$,
and the vertical principal point at the image center. We project
$\mathbf{K}^{\mathsf T}\mathbf{F}_{ij}\mathbf{K}$ onto the
essential-matrix manifold and evaluate the mean Sampson cost
$\mathcal{C}_{ij}(f,c_x)$. The refined focal is
\begin{equation}
    f_{\mathrm{ref}}
    =
    \arg\min_{f\in\mathcal{F}}
    \min_{c_x\in\mathcal{C}}
    \frac{1}{|\mathcal{R}|}
    \sum_{(i,j)\in\mathcal{R}}
    \mathcal{C}_{ij}(f,c_x),
    \label{eq:rgb_revisit_refinement}
\end{equation}
where $\mathcal{F}$ and $\mathcal{C}$ are the focal and horizontal-center
search ranges specified below. We profile over $c_x$ as a nuisance
parameter and update only the focal, retaining the centered downstream
camera model. Different revisit headings supply complementary epipolar
constraints from RGB correspondences.

\paragraph{Routing and configuration.}
Refinement requires sufficient revisits and continued focal
evolution late in base calibration. We sample every five frames,
propose pairs within 15 m and at least 200 frames apart, and require at least
1,000 proposals, a density of at least 0.15 per frame, and a late focal
change of 0.2--2.5\%. At most 800 pairs with heading changes of
$25^\circ$--$140^\circ$ are retained. Given the base focal $f_0$, we
search 401 log-spaced focal values in $[0.6f_0,1.3f_0]$ and 41 horizontal
centers within $\pm40$ pixels of the image midpoint, with 60
matches per pair and 400 bootstrap resamples.

We accept only interior, bootstrap-stable profile minima, allowing at most
10\% boundary solutions, 15\% disagreement between center-pinned and
center-profiled fits, and 25\% focal change from $f_0$. A rejected fit may
be retried once with upsampled matching under the same checks; otherwise,
the base calibration is retained. Pair construction is deterministic, and
routing and acceptance use no sensor intrinsics, ground-truth poses, or
ground-truth depths. Refinement averages 7.1 minutes per routed sequence,
dominated by matching; the focal-profile fit takes 4--6 seconds.

\begin{table*}[t]
\centering
\caption{\textbf{Effect of RGB-only focal refinement on \textbf{VBR}.} We report
full-sequence ATE (m; lower is better) with and without loop closure.
The local estimator is held fixed.}
\label{tab:vbr_calibration_stage_b}
\scriptsize
\setlength{\tabcolsep}{4.2pt}
\resizebox{\textwidth}{!}{%
\begin{tabular}{l*{7}{c}c}
\toprule
\textbf{Variant} &
\textbf{colosseo\_0} &
\textbf{campus\_0} &
\textbf{campus\_1} &
\textbf{pincio\_0} &
\textbf{spagna\_0} &
\textbf{diag\_0} &
\textbf{ciampino\_1} &
\textbf{Avg.} $\downarrow$ \\
\midrule
Ours w/o refinement &
30.73 & 45.59 & 14.41 & 15.78 & 18.65 & 35.35 & 74.07 & 33.51 \\
Ours &
30.08 & 9.26 & 15.07 & 7.43 & 13.94 & 20.97 & 17.75 &
\textbf{16.36} \\
\midrule\midrule
Ours w/o refinement w/ LC &
23.44 & 37.27 & 9.49 & 14.46 & 13.36 & 34.80 & 62.85 & 27.95 \\
Ours w/ LC &
26.83 & 6.80 & 9.99 & 6.94 & 10.18 & 5.84 & 9.60 &
\textbf{10.88} \\
\bottomrule
\end{tabular}}
\end{table*}

\paragraph{VBR results.}
Under matched inference settings, refinement reduces mean ATE from 33.51
to \textbf{16.36 m} without loop closure (51\%) and from 27.95 to
\textbf{10.88 m} with loop closure (61\%)
(\cref{tab:vbr_calibration_stage_b}).

\begin{table*}[t]
\centering
\caption{\textbf{Retrospective focal analysis on \textbf{VBR} across ten independent
runs.} Entries are signed relative focal errors against sensor calibration,
used only for reporting, never for routing, refinement, or model selection.
Macro averages report absolute errors.}
\label{tab:vbr_calibration_focal}
\scriptsize
\setlength{\tabcolsep}{4.0pt}
\resizebox{\textwidth}{!}{%
\begin{tabular}{l*{7}{c}c}
\toprule
\textbf{Variant} &
\textbf{colosseo\_0} &
\textbf{campus\_0} &
\textbf{campus\_1} &
\textbf{pincio\_0} &
\textbf{spagna\_0} &
\textbf{diag\_0} &
\textbf{ciampino\_1} &
\makecell{\textbf{Macro}\\\textbf{abs. error}} \\
\midrule
Ours w/o refinement & +7.68\% & +8.26\% & +1.25\% & +8.04\% &
+7.18\% & +8.78\% & +8.37\% & 7.08\% \\
Ours & \textbf{+5.56\%} & \textbf{+1.33\%} & +1.25\% &
\textbf{-1.41\%} & \textbf{+4.89\%} & \textbf{-0.06\%} &
\textbf{+0.14\%} & \textbf{2.09\%} \\
\bottomrule
\end{tabular}}
\end{table*}

Across ten runs, macro absolute focal error decreases from 7.08\% to
\textbf{2.09\%} (\cref{tab:vbr_calibration_focal}); sensor values
are used only for retrospective evaluation. Trajectory rows use independent
runs, so small ATE differences can occur even without a focal update.

\begin{table*}[t]
\centering
\caption{\textbf{Effect of camera-model choice on \textbf{VBR}.} We report full-sequence
ATE (m; lower is better) with and without loop closure.
The local estimator is held fixed within each comparison.}
\label{tab:vbr_calibration_gt_k}
\scriptsize
\setlength{\tabcolsep}{4.2pt}
\resizebox{\textwidth}{!}{%
\begin{tabular}{l*{7}{c}c}
\toprule
\textbf{Variant} &
\textbf{colosseo\_0} &
\textbf{campus\_0} &
\textbf{campus\_1} &
\textbf{pincio\_0} &
\textbf{spagna\_0} &
\textbf{diag\_0} &
\textbf{ciampino\_1} &
\textbf{Avg.} $\downarrow$ \\
\midrule
Ours (Calibrated) &
18.19 & 16.13 & 15.78 & 6.34 & 14.35 & 22.77 & 41.64 & 19.31 \\
Ours (Uncalibrated)&
28.38 & 8.46 & 15.42 & 7.73 & 13.42 & 21.70 & 18.77 &
\textbf{16.27} \\
\midrule\midrule
Ours (Calibrated) w/ LC &
12.92 & 13.06 & 7.46 & 5.50 & 12.68 & 8.34 & 25.82 & 12.25 \\
Ours (Uncalibrated) w/ LC &
21.54 & 6.02 & 9.75 & 7.18 & 9.55 & 5.53 & 13.06 &
\textbf{10.38} \\
\bottomrule
\end{tabular}}
\end{table*}

\cref{tab:vbr_calibration_gt_k} compares the RGB-estimated camera
model with supplied sensor intrinsics. Average ATE decreases from 19.31 to
\textbf{16.27 m} without loop closure and from 12.25 to
\textbf{10.38 m} with loop closure. Sensor intrinsics provide the full
off-center matrix to geometric optimization, while depth prediction uses
only horizontal field of view. The RGB-only variant instead estimates an
effective focal for a centered camera. Its lower ATE reflects compatibility
within this composed camera-and-depth model, not a more accurate estimate of
the physical camera matrix.

\subsection{Implementation Details}
\label{sec:implementation}

\paragraph{Training configuration.}
We initialize the local estimator from the official DPVO checkpoint and train
on 16 NVIDIA A100 80GB GPUs with distributed data parallelism. Each GPU
processes one 15-frame clip, giving a global batch size of 16. Training spans
15,000 optimizer steps, or 240,000 clips. We first warm up the
zero-initialized representation-fusion branches on 5,000 TartanAir clips
while keeping the odometry network frozen. We then jointly fine-tune the
odometry network and fusion branches using the remaining 235,000 clips
from our multi-dataset training mixture of TartanAir,
TartanGround, CarlaOcc,
LightwheelOcc, and
OmniWorld-Game. The resulting training exposure is
56,000 TartanAir clips and 46,000 clips from each of the other four datasets.
No training clip exceeds 15 frames, and no evaluation sequence or dedicated
long-sequence pose corpus is used.

\paragraph{Optimization and precision.}
We use a frozen DINOv3-S/16 encoder pretrained on LVD-1689M
\citep{simeoni2025dinov3}. Its 384-dimensional features are projected to 64
channels and split into two 32-channel branches, which augment the
128-dimensional matching stream and the 384-dimensional recurrent context
stream, respectively. The final residual projections are zero initialized.
In joint training, the learning rates are $8\times10^{-6}$
for the odometry network and $8\times10^{-5}$ for the fusion branches. We use
AdamW with weight decay $10^{-6}$, linear OneCycle annealing with a 1\% warm-up,
gradient clipping at 10, and seed 1234. For each clip, we sample 80 $3\times3$ patches
per frame and unroll 18 recurrent refinements. Images are resized while
preserving aspect ratio and center-cropped to $480\times640$. Only the frozen
DINO forward pass uses BF16, evaluated in three-frame chunks; the adapters,
odometry network, differentiable bundle adjustment, and losses remain in FP32.
MoGe-2 is not used during adaptation and is kept frozen when invoked at
inference. End-to-end training takes approximately seven hours.

\paragraph{Metric-prior inference.}
DPVO and MoGe-2 receive the same cover-fitted and center-cropped image raster,
and we transform the camera intrinsics by the same resize and crop. No
additional warp is applied to the predicted depth. For the canonical KITTI
input, an image is mapped to $280\times504$ pixels. Writing the full-resolution
output as $\widehat{\mathbf{D}}^{\mathrm{full}}_i$, the registered depth used
in Sec.~\ref{sec:metric_prior} is
$\widehat{\mathbf{D}}_i=\widehat{\mathbf{D}}^{\mathrm{full}}_i[1::4,1::4]$;
a patch center $(x,y)$ thus maps to image pixel
$(4x+1,4y+1)$.

A predicted depth is valid when the provider mask is true, its value is finite,
and $0.2\,\mathrm{m}\leq z\leq500\,\mathrm{m}$, including both endpoints. An
invalid patch is initialized with the median valid inverse depth of the full
registered frame and receives zero metric-factor weight. Metric inference
samples 96 patch centers per frame, whereas training uses 80. If the
full-resolution prediction, the stride-4 depth raster, or all sampled patch
centers contain no valid depth, the current sequence is terminated and no
partial trajectory is reported as a valid result.

We disable DPVO's scale normalization in every metric run. The metric factor
uses $\lambda_{\mathrm{m}}=1$ and Huber threshold $\delta=0.3$. It is disabled
for the first 12 accepted graph frames and latched thereafter, so the first
regular BA update with this factor normally occurs on accepted frame 13.

\paragraph{DINOv3 feature alignment.}
Before DINOv3 inference, we undo DPVO's input normalization, convert BGR to RGB,
and replicate-pad the right and bottom boundaries to the next multiple of 16.
This padding is confined to the DINOv3 branch and does not change DPVO image
coordinates or camera intrinsics. Frozen DINOv3-S/16 produces a stride-16
raster of 384-dimensional tokens. A $1\times1$ convolution projects these
tokens to 64 channels, split into 32-channel matching and context branches.
We align the branch features with DPVO's grid by bilinear upsampling to stride
4 in the padded coordinate system and then remove the
padded rows and columns. For a $280\times504$ input, the alignment is
$288\times512\rightarrow18\times32\rightarrow72\times128\rightarrow70\times126$.
The matching branch projects to 128 channels and is added to both source and
target full-frame matching maps before patch extraction and correlation.
The context branch is sampled at source patch centers, projected to 384
dimensions, and added to the corresponding recurrent context vectors.
The final residual projections are initialized to zero.

\paragraph{Loop closure.}
\label{sec:appendix_loop_closure}
Following VGGT-Long, Scal3R, and
HorizonStream, we use a similar loop-closure
strategy that combines revisit retrieval with graph refinement.
We use SALAD to retrieve temporally separated
near-duplicate views and suppress redundant matches. Each selected pair
contributes a weak identity relative-pose factor, treating the views as
approximately coincident without pairwise geometric verification.
These factors are combined with sequential odometry constraints in an
$\mathrm{SE}(3)$ pose graph, which refines camera poses with the first pose
fixed. This correction preserves metric scale and leaves patch depths
unchanged; dense reconstructions can be reassembled from the corrected poses.

\paragraph{Stereo inference.}
For each newly created left-image patch, we add a same-time edge to the
synchronized right image under the fixed left-to-right rig transform. The right
image supplies only a target matching map, including its DINOv3 matching
augmentation; no right-image patch, recurrent context, metric-depth variable,
or camera-pose variable is created. The recurrent matcher predicts the right
correspondence and confidence, and the resulting reprojection residual is added
to the local BA objective. No stereo-specific data or loss is used during
training.

\end{document}